\documentclass[10pt,final,journal]{IEEEtran}
\usepackage[ruled, vlined]{algorithm2e}
\usepackage{amsmath, amssymb, amsthm, amsbsy}
\usepackage{xcolor}
\usepackage{bbm}
\usepackage{tcolorbox}
\usepackage{tikz}
\usepackage{float}
\usepackage{bm}
\usepackage{autobreak}
\newcommand{\xiRatio}{ {\frac{ {\hat \Delta}_{k|k-1;i;j}^U}{ {\hat r}_{k|k-1;i;j}}}}

\newcommand{\psiRatio} { {\frac{ {\hat \Delta}_{k|k-1;i;j}^U}{ {\hat r}_{k|k-1;i;j}^2}}}
\newcommand{\betaZRatio}  { {\frac{ {\hat \Delta}_{k|k-1;i;j}^U}{ {\hat r}_{k|k-1;i;j}}}}
\newcommand{\DeltaUp} { {\hat \Delta}_{k|k-1;i;j}^U}
\newcommand{\rHat} { {\hat r}_{k|k-1;i;j}}

\begin{document}

\newtheorem{thm}{Theorem}
\newtheorem{rslt}{Result}

\title{ Three-Dimensional Swarming Using Cyclic Stochastic Optimization}
\author{Carsten H. Botts \\ The Johns Hopkins University Applied Physics Lab \\ Laurel, MD \\ Email: Carsten.Botts@jhuapl.edu}

\maketitle

\begin{abstract} In this paper we simulate an ensemble of cooperating, mobile sensing agents that implement the cyclic stochastic optimization (CSO) algorithm in an attempt to survey, track, and follow multiple targets.    In the CSO algorithm proposed, each agent uses its sensed measurements,  its shared information, and its predictions of other agents' future motion to decide on its next action.   This decision is selected to minimize a loss function that decreases as the uncertainty in the  target state estimates decreases.    Only noisy measurements of this loss function are available to each agent, and in this study, each agent attempts to minimize this function by calculating its gradient. This paper examines, via simulation-based experiments,  the implications and applicability of CSO convergence  in three dimensions.   \end{abstract}


\section{Introduction}
\label{sec:intro}  


This paper examines stochastic decentralized resource optimization in the context of a multi-agent, multi-target surveillance mission.  The resources we consider are mobile/unmanned agents which are capable of selecting their own motion. The setting of this problem is deliberately generic, so these unmanned agents could, for example, be aerial or underwater. The mission of these agents is to learn as much as possible about the kinetic states  (position and velocity) of the  nearby targets. They do this by iteratively (one agent at-a-time) minimizing a loss function which is stochastically measured. This loss function decreases with growing information of the targets, and by minimizing this loss function, the agents end up tracking the targets. This idea was initially explored in \cite{Peterson1} and \cite{Botts1}, but these studies were done in two dimensions. This paper builds on those results by extending the study to three dimensions, implementing a more accurate and stable estimation method, and considering the applicability of this swarming algorithm for a large number of agents.   We even investigate how multiple groups of agents (each group containing several agents and working independently of any other group) track and follow several targets.

 This  section begins by giving details of the problem, discussing the assumptions we make, and comparing these aspects of our study with those that are currently in the swarming literature. This is done in Section \ref{sec:intro_probDescription}.      Some brief details regarding the loss function are then given in Section \ref{sec:intro_lossFunction}.    In Section \ref{sec:Centralized_vs_Decentralized}, we discuss the various methods that can be used to minimize such a loss function.


\subsection{Problem Description and Assumptions}
\label{sec:intro_probDescription}


The  problem discussed in this paper is how to configure several mobile sensing agents over time and 3D space so that their awareness of the targets is optimized.   We say that an agent's ``awareness'' of a target increases if the uncertainty in its estimates of the target's states decreases.      

We assume that some aspects of the agents and targets are unpredictable in time, including target motion and sensing reliability. Since aspects such as these change with time, no steady-state solution exists;  any optimal solution at a particular time may not be optimal in the immediate future.   

The agents optimize their configuration and orientation using decentralized motion planning. In decentralized motion planning, each agent decides on its heading and vertical displacement at each time step via minimization of a loss function.   For each agent, this minimization is attempted analytically, i.e., it is done in one step and it is done by taking the gradient of the stochastic loss function. These are the two primary features of our proposed algorithm that distinguish it from other swarming-type algorithms: (1) the loss function we are minimizing changes at each time step, which necessitates a quick (analytical and one-step) solution, and  (2) the loss function being minimized is stochastic. The efficacy of other swarming algorithms in optimizing a stochastic function is still not clear, while certain studies have shown that CSO does converge when optimizing noisy functions (see  Ref. \cite{Hernandez}).  Many of the most common  swarming optimization algorithms are also derivative-free and require multiple steps in optimizing a single function. Such algorithms would be impractical in the setting we consider, and some of these algorithms are reviewed below.

  Genetic algorithms are swarming optimization algorithms which  begin by considering a set, or an initial population, of possible solutions.  The feasibility or fitness of each solution in this set is then examined, and the set is then accordingly refined into a new set of possible solutions.  This new set can be thought of as the next generation of the population. This refinement is done using an action that is modeled after genetic/evolutionary processes, such as  mutation, crossover, or reproduction.  Genetic algorithms continue to refine this set/population  until they find the optimal solution.  Genetic algorithms have been successfully applied to many problems including those in operations management (see Ref.  \cite{Lee}) and health (see Refs. \cite{Reddy, Devarriya}).   Differential evolution is another swarming optimization algorithm similar to genetic algorithms. It too begins with an initial population, or set of possible solutions. From these solutions, a new set of candidate solutions are created, and if a candidate is superior to its parents, it replaces its parents in the set of solutions. Variations of it have also been successfully applied to many problems (see Refs. \cite{Chai, Sickel, Jebaraj}).    The Artificial Bee Colony (ABC) algorithm  (see Refs. \cite{Karaboga, KarabogaAkay}) is a swarming optimization algorithm inspired by the food-seeking behavior of honey bees. In the ABC algorithm, the optimal solution is found by  ``employed" bees  exploring solutions/sources of food, communicating their findings to ``onlooking" bees, after which the onlooking bees select a source of food (a solution) that is better than the current one.    Particle Swarm Optimization (PSO) is yet another swarming optimization algorithm modeled after animal behavior (bird flocking).  Like the other swarming algorithms, PSO begins with a set of candidate solutions/particles. These particles  swarm towards the optimal solution by iteratively evaluating the feasibility of the particles in the set, and flocking towards those particles which have good (or better) solutions.  Just as with genetic algorithms and differential evolution, ABC and PSO have been successfully applied to solve a myriad of problems (see Refs \cite{Mac, Bansal, Jadhav}).  In some cases, different types of swarming algorithms are combined into one overall optimization algorithm. In Fares et al. (Ref. \cite{Fares}), for example, a recently developed swarming algorithm called the whale optimization algorithm (see Ref. \cite{Mirjalili}) is initially used to explore spaces of the function to be optimized.   The PSO then acts on the feedback from the whale optimization algorithm to finally optimize the function.

Even if convergence of these swarming algorithms in the presence of a noisy objective function was guaranteed, applying algorithms such as these to our problem would be impractical. In our case, the agents are not swarming to the same point that minimizes the loss function. They are seeking the optimal spatial arrangement, and they are doing this at each time point. To apply any of the algorithms mentioned above, multiple configurations of the agents would have to be considered  and repeatedly acted upon to find the optimal one.   This procedure would also have to happen at each time point since the loss function at each time point is different.  A more efficient strategy is preferred, and that  is why we consider minimizing the loss function as we do.   This loss function is discussed in the section below.


\subsection{The Loss Function}
\label{sec:intro_lossFunction}


  As in \cite{Botts1},  the loss function  is information-based, stochastic, and time-varying.  By information-based, we mean that the function to be minimized quantifies the expected information gain resulting from agents making specific motions. Having agents or sensors select motions or actions to maximize some measure of information on targets has been done before.  In Sinha et al. (Ref. \cite{Sinha}), for example, UAVs make decisions to optimize an objective function which optimizes the detectability and information on the kinetic states of ground-based targets.  Kreucher et al. (Ref. \cite{Kreucher}) has sensors select actions (where to move and what direction to emit energy, for example) to maximize information gain on targets, and Yang et al. (Ref. \cite{Yang}) even studied sensor resource management when information gain is optimized yet defined in alternative ways.  
  
  The loss function we employ is also stochastic since it is a function of the agent's measurements of the targets. These measurements are random, making the loss function random.   Randomness (or stochasticity) in the loss function changes the minimization process because the algorithm often gets misleading information about the fitness of the solution.

Finally, the loss function is time-varying since the agents and targets move at every time step, and the motion of the targets may be entirely unpredictable. As the agents and targets move, the detectability of the targets, the view geometries of the agents, and the agents' ability to communicate change. These changes affect the loss function.     The subsection below discusses two ways in which the loss function can be minimized.


\subsection{Centralized vs. Decentralized Optimization}
\label{sec:Centralized_vs_Decentralized}


Centralized and decentralized optimization methods each have a role in resource optimization. In centralized optimization,  the action of all agents are optimized simultaneously, whereas in decentralized optimization, each agent selects its optimal move separately. 

Because our agents do not perfectly communicate, they can not always share information, and their decisions are often based on limited data. Each agent must thus act individually.  If communications were not a problem, all agents would simultaneously have the exact same information of all the targets, and all the agents would coordinate together to arrive at a globally optimal solution. This globally optimal solution is identical to centralized optimization.

Some literature has recently emerged on cooperative control of multiple agent systems  (see Ref.~\cite{Gazi} for a current and broad overview).  Tang et al. (Ref. \cite{Tang}), for example, studied how agents can cooperate to minimize the average amount of time between target detections.  Jin et al. (Ref. \cite{Jin}) investigated how effectively agents cooperate if the number and location of the targets are known {\it a priori}, and  DeSena et al. (Ref. \cite{DeSena13}) studied the effectiveness of multi-agent decentralized collaboration as a function of communication connectivity.  It is still unclear, however, exactly how the decentralized optimization process degrades the centralized solution when the function being optimized is stochastically measured.   Some preliminary theoretical results have recently emerged, though, regarding convergence conditions on the cyclic optimization of stochastic functions (see Ref. \cite{Hernandez}).

   In Section \ref{sec:joint-estim-ctrl-prob}, more details of the decentralized control problem  are described.  In Section  \ref{sec:EKF}, we describe how each agent estimates the states of the targets, and in Sections  \ref{sec-lossFunctionFormulation} and \ref{sec:cso-seesaw}, we describe the details of how an agent minimizes this loss function.   In Section \ref{sec:simulations}, the results of some simulations are given, and in Section \ref{sec:conclusions}, conclusions and areas of future work are discussed.


\section{Decentralized Control Problem Definition}
\label{sec:joint-estim-ctrl-prob}  

\begin{figure}[th]
   \begin{center}
   \begin{tabular}{c}
      \includegraphics[width=0.4\textwidth]{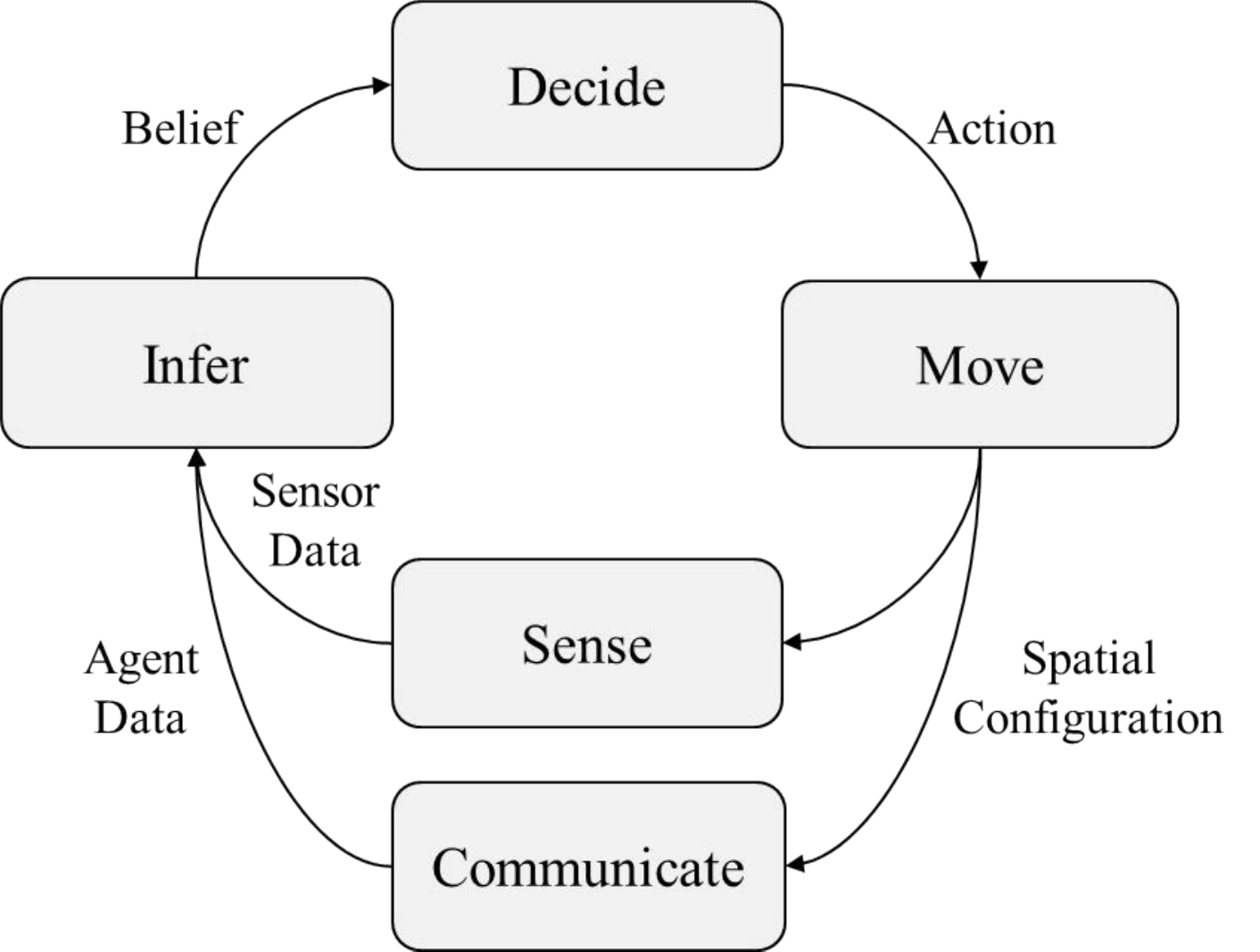}
   \end{tabular}
   \end{center}
   \caption[example] 
   { \label{fig:per-agent-planning-cycle} Per-agent perception-action planning cycle. }
\end{figure}

  We have a decentralized control problem in which each agent decides its actions according to its  estimates of targets and its information communicated to it by other agents.  We refer to each agent's iterative estimation and control process as its \emph{perception-action cycle} (PAC) (see, e.g., Ref.~\cite{Haykin}).  For the agents considered in this paper, the PAC  consists of the following steps: sense, communicate,  infer, decide, and move.  This process is illustrated in Figure~\ref{fig:per-agent-planning-cycle}, and a description of the steps follows.  In the description, the true state of  target $i$ at time $k$ will be denoted as $${\bf x}_{k;i} = \left( x_{k;i}^E, x_{k;i}^N, x_{k;i}^U,  \dot{x}_{k;i}^E, \dot{x}_{k;i}^N, \dot{x}_{k;i}^U \right),$$ where $x_{k;i}^E$ is the east coordinate of target $i$ at time $k$, $x_{k;i}^N$ is its north coordinate, $x_{k;i}^U$ is its up (or vertical) coordinate, ${\dot{x}}_{k;i}^E$ is its velocity in the east direction, ${\dot{x}}_{k;i}^N$ is its velocity in the north direction, and ${\dot{x}}_{k;i}^U$ is its velocity in the up direction. And in the simulations we conducted in this study, the procedure described below was executed for each agent at each time step.

\begin{enumerate}
\item SENSE: The agent measures the range, azimuth angle and polar angle to all the targets that it senses.   The range, azimuth angle and polar angle that agent $j$ measures to target $i$  at time $k$ will be denoted  $r_{k;i;j}$, $\phi_{k;i;j}$ and $\theta_{k;i;j}$, respectively, where   \begin{eqnarray*}   r_{k;i;j} &  = &  \sqrt{( \Delta_{k;i;j}^E )^2 + (\Delta_{k;i;j}^N)^2+ (\Delta_{k;i;j}^U)^2}  \\  \phi_{k;i;j} &  = & \tan^{-1}  \left[  (\Delta_{k;i;j}^N ) / ( \Delta_{k;i;j}^E )  \right],  \\ \theta_{k;i;j} & = &  \cos^{-1} \left[  \left. \left(  \Delta_{k;i;j}^U  \right) \right/ r_{k;i;j} \right], \\ \Delta_{k;i;j}^E & = & x_{k;i}^E - y_{k;j}^E, \\ \Delta_{k;i;j}^N & = & x_{k;i}^N - y_{k;j}^N, \\ \Delta_{k;i;j}^U & = & x_{k;i}^U - y_{k;j}^U,   \end{eqnarray*}    and $ \left( y_{k;j}^N, y_{k;j}^E, y_{k;j}^U \right)$ are the north, east, and up coordinates of agent $j$.
   
\item COMMUNICATE: The agent sends state estimates, Fisher information matrices, and its most recent motion decision to peer agents.  The reliability and latency of these communications vary with the type of vehicle considered and the environment in which they are operating.  Pantelimon et al. (Ref. \cite{Pantelimon}) review the communication strategies and hardware involved in various types of unmanned vehicle deployments. For aerial vehicles, Wi-Fi modules are the most common hardware. They have a communication range close to 100m and a communication latency on the order of one millisecond.  The size and the necessary programming involved in setting up such Wi-Fi hardware is a drawback, however. Bluetooth is a less reliable alternative and has a smaller communication range, but it may be preferable to Wi-Fi as it is lower in weight and complexity.  Acoustic communications are the best for underwater vehicles, but the weight, complexity and cost of the hardware involved in such communications is relatively high. The latency is also quite large (close to 0.67 ms/m, see Ref. \cite{Burrowes}). The simulations we conduct in this study  incorporate a model for gradual attenuation of communication over distance, and this is consistent with real-world performance.   

\item INFER: Given the information sensed, the agent updates the state estimate for each detected target via  a second order extended Kalman filter.    The state estimate agent $j$ has of target $i$ at time $k$ given all the data up until (and including) time $k$ will be denoted as  \begin{eqnarray*} {\hat  {\bf x}}_{k|k;i; j}  &  = & \left( {\hat {x}}_{k|k;i;j}^E, {\hat{ x}}_{k|k;i;j}^N, {\hat{ x}}_{k|k;i;j}^U, \right.  \\  & &~~ \left.  {\widehat {\dot{x}}}_{k|k;i;j}^E, {\widehat {\dot{x}}}_{k|k;i;j}^N, {\widehat {\dot{x}}}_{k|k;i;j}^U \right). \end{eqnarray*} The state estimate agent $j$ has of target $i$ at time $k$ given all the data up until (but not including) time $k$ will be denoted as \begin{eqnarray*} {\hat  {\bf x}}_{k|k-1;i; j}&  = & \left( {\hat {x}}_{k|k-1;i;j}^E, {\hat{ x}}_{k|k-1;i;j}^N, {\hat{ x}}_{k|k-1;i;j}^U, \right.  \\ & &~~ \left. {\widehat {\dot{x}}}_{k|k-1;i;j}^E, {\widehat {\dot{x}}}_{k|k-1;i;j}^N, {\widehat {\dot{x}}}_{k|k-1;i;j}^U \right). \end{eqnarray*}  With these estimates and the other information communicated to it from peer agents, the agent estimates the loss function. The details of this loss function are given in Section \ref{sec-lossFunctionFormulation}.
   
\item DECIDE: The agent selects its next action (its heading and vertical displacement) by minimizing its estimated loss function.   The details of this step  are given in Section \ref{sec:cso-seesaw}.
\item MOVE: Each agent updates its state according to its selected action (its selected heading and vertical displacement).  It is assumed that the time scale is large enough that rotational dynamics are negligible, and the vehicle can instantaneously change direction.  In this paper, the state of agent $j$ at time $k$ will be denoted as  ${\bf y}_{k;j} = \left( y_{k;j}^E, y_{k;j}^N, y_{k;j}^U, \dot{y}_{k;j}^E, \dot{y}_{k;j}^N, \dot{y}_{k;j}^U \right)$, and the heading of  agent $j$  at time $k$ will be denoted $\gamma_{k;j}$.   The relation between $\gamma_{k;j}$ and ${\bf y}_{k;j}$ is $\gamma_{k;j} = \tan^{-1} \left(   \left. \dot{y}_{k;j}^N \right/  \dot{y}_{k;j}^E  \right)$.   

\end{enumerate} 

As mentioned earlier, these steps are meant to occur sequentially in agents and among agents, implying that an agent will execute the entirety of these steps, then another agent will, then another will, etc.  The time difference between agents executing the PAC steps thus has to account for communication latency and the time it takes an agent to do the necessary processing and computing. 

Section \ref{sec:EKF} gives the details on Steps 2 - 4, i.e., it discusses how the agents sense and estimate the state of the targets.  Section \ref{sec-lossFunctionFormulation} then gives details on how each agent estimates and attempts to minimize the loss function (Steps 3-4).

\section{Estimating the State of the Target}
\label{sec:EKF}

This section describes how an agent estimates the states of a target that it senses.

We begin by letting $S_{k;i;j} = 1$ if agent $j$ senses target $i$ and $S_{k;i;j} = 0$ otherwise. Each agent also assumes that the motion of target $i$ can be characterized with the state equation \begin{equation} \label{eqn:TrgtMotion} {\bf x}_{k;i} = \Phi {\bf x}_{k-1; i}  + {\bf w}_{k;i}, \end{equation} where ${\bf w}_{k;i} \sim N \left( {\bf 0}_{6 \times 1}, {\bf Q} \right),$ and \begin{equation} \label{eqn:PhiMatrix} \Phi = \left( \begin{array}{cccccc} 1 & 0 & 0 &  \Delta t & 0 & 0 \\ 0 & 1 & 0 & 0 & \Delta t & 0 \\ 0 & 0 & 1 & 0 & 0 & \Delta t \\ 0 & 0 & 0 & 1 & 0 & 0 \\ 0 & 0 & 0 & 0 & 1 & 0 \\ 0 & 0 & 0 & 0 & 0 & 1 \end{array} \right). \end{equation}  The coordinates in Equation \ref{eqn:TrgtMotion} are in East, North, and Up, yet the agents sense range, azimuth and polar angle. The measurements agent $j$ makes of target $i$ are thus $$ {\bf z}_{k;i;j} = h \left( {\bf x}_{k;i} \right) + {\bf v}_{k;i;j} = \left( r_{k;i;j}, \phi_{k;i;j}, \theta_{k;i;j} \right)^T + {\bf v}_{k;i;j},$$ where ${\bf v}_{k;i;j} \sim N \left( {\bf 0}_{3 \times 1}, {\bf R} \right),$ and ${\bf R} = {\rm diag} \left( \sigma^2_r, \sigma^2_\phi, \sigma^2_{\theta} \right).$

The agent then applies a second order  extended Kalman filter to obtain estimates of ${\bf x}_{k;i}$. A second order extended Kalman filter is used rather than a first order extended Kalman filter since it produces  more accurate state estimates and the additional computational cost of using it is negligible.  The details of how agent $j$ calculates these estimates of target $i$ are given  in the text box labeled ``The Second Order Extended Kalman Filter."

 With the estimates calculated in the extended Kalman filter, an agent estimates a loss function. The motion it takes at time $k$ is selected to minimize this loss function.   The details of the loss function and its minimization are given in the section below.

\section{Formulation and Minimization of the Loss Function}
\label{sec-lossFunctionFormulation}


 As mentioned earlier, in each DECIDE step an agent selects a heading and vertical displacement. This is done  to minimize a loss function which measures information over the targets. The more information on the targets, the smaller the loss function.     In performing its local minimization, however, each agent can only approximate the  loss function using estimates of target states and predictions of peer agent actions. 

We define the ``information'' about target $i$  using the Fisher information matrix (FIM) of the state estimates on target $i$.   We denote the total pre-action FIM of target $i$ at time $k$  as ${\bf F}_{k|k-1;i}^{\rm Total}$, where ${\bf F}_{k|k-1;i}^{\rm Total}$ is the sum of every agent's knowledge (or ``information'') about target $i$ before data at time $k$ has been processed.  It is meant to measure the entire information of target $i$ before time $k$, and it is mathematically written as  \begin{equation} \label{eqn:preactionFisher}  {\bf F}_{k|k-1;i}^{\rm Total} = \sum_{j=1}^A {\bf F}_{k|k-1;i;j}, \end{equation}  where $A$ is the total number of agents, and ${\bf F}_{k|k-1;i;j}$ is the pre-action FIM that agent $j$ has on target $i$'s states at time $k$, i.e., ${\bf F}_{k|k-1;i;j} = {\bf P}_{k|k-1;i;j}^{-1}$.  The total pre-action FIM of target $i$ at time $k$ (as calculated in  (\ref{eqn:preactionFisher})) would be equivalent to the true total Fisher information on target $i$ if all measurements and estimates on the target were independent, which they are not. Independence of these measurements and estimates is an assumption we are making in this study.

\begin{tcolorbox}

\begin{center}
\begin{large}

The Second Order Extended Kalman Filter

\end{large}
\end{center}

Let ${\hat {\bf x}}_{{k|k-1};i;j}$ be the predicted state estimate of target $i$ by agent $j$, and let ${\bf P}_{k|k-1;i;j}$ be the corresponding predicted error covariance of ${\hat {\bf x}}_{{k|k-1};i;j}$.    That is, $${\bf P}_{k|k-1;i;j} = {\mathbbm E} \left[  \left(  {\hat {\bf x}}_{{k|k-1};i;j} - {\bf x}_{{k};i}  \right) \left(   {\hat {\bf x}}_{{k|k-1};i;j} - {\bf x}_{k;i}   \right)^T   \right].$$   Then$${\hat {\bf x}}_{k|k;i;j} = {\hat {\bf x}}_{{k|k-1};i;j} + {\bf K}_{k;i;j} {\bf u}_{k;i;j} {\mathbbm 1} \left( S_{k;i;j} = 1 \right),$$  and $${\bf P}_{k|k;i;j} = \left[ {\bf I}_{6 \times 6} - {\bf K}_{k;i;j} {\bf H}_{k;i;j} {\mathbbm 1} \left( S_{k;i;j} = 0 \right) \right] {\bf P}_{k|k-1;i;j},$$ where   \begin{eqnarray*}    {\bf K}_{k;i;j} & = & {\bf P}_{k|k-1;i;j} {\bf H}_{k;i;j}^T {\bf S}_{k;i;j}^{-1},   \\ {\bf H}_{k;i;j} & = &  \left. {\frac{\partial h}{\partial x}}  \right|_{{\bf x} = {\hat {\bf x}}_{k|k-1;i;j}}, \\ {\bf S}_{k;i;j}[l,m] & = & {\bf H}_{k;i;j} {\bf P}_{k|k-1;i;j}  {\bf H}_{k;i;j}^T + {\bf R} + \\ & & \left[ {\frac{1}{2}} {\rm tr} \left( \nabla_x^2 h_l \left(   {\hat {\bf x}}_{k|k-1;i;j} \right){\bf P}_{k|k-1;i;j}  \cdot \right. \right. \\ & & \left. \left.    \nabla_x^2 h_m \left(   {\hat {\bf x}}_{k|k-1;i;j} \right){\bf P}_{k|k-1;i;j} \right) \right]   \\ {\bf u}_{{k};i;j}[l] & = & {\bf z}_{{k};i;j} - \left \{ h \left( {\hat {\bf x}}_{{k|k-1};i;j} \right) +  \right. \\ & &  \left.  {\frac{1}{2}} {\rm tr} \left[  \nabla_x^2 h_l \left(   {\hat {\bf x}}_{k|k-1;i;j} \right){\bf P}_{k|k-1;i;j}  \right]  \right \}. \end{eqnarray*}

  With ${\hat {\bf x}}_{k|k;i;j}$ it is straight-forward for agent $j$ to obtain the predicted value of ${\bf x}$ for target $i$ at time $k+1$.   This is ${\hat {\bf x}}_{k+1|k;i;j} = {\pmb \Phi} {\hat {\bf x}}_{k|k;i;j}$  and the predicted error covariance associated with this estimate is ${\bf P}_{k+1|k;i;j} = \Phi {\bf P}_{k|k;i;j} \Phi^T + {\bf Q}$. The detailed calculations of ${\bf H}_{k;i;j}$ are given in the Appendix.

\end{tcolorbox}

We specifically use the log determinant of the pre-action Fisher information, $\log | {\bf F}_{k|k-1;i}^{\rm Total} |$, to quantify information about target $i$ before new data is acquired, and the log determinant of the post-action Fisher information, $\log | {\bf F}_{k|k;i}^{\rm Total}({\pmb \xi}_k^{\rm all}) |$, to quantify  information about target $i$   after new data is acquired. This new data is obtained  as a consequence of all the agents executing actions ${\pmb \xi}_k^{\rm all}$ at time $k$, where  ${\pmb \xi}_k^{\rm all} = \left( {\pmb \xi}_{k;1}, {\pmb \xi}_{k;2}, \ldots, {\pmb \xi}_{k;A} \right)$, ${\pmb \xi}_{k;j}$ is the action taken at time $k$ by agent $j$, ${\pmb \xi}_{k;j} = \left( \gamma_{k;j}, y^U_{k;j} \right)^T,$  $\gamma_{k;j}$ is the bearing agent $j$ takes at time $k$, and $y^U_{k;j}$ is the vertical position it chooses at time $k$.  Mathematically, $\gamma_{k;j}$  can be written as \begin{equation*}  \gamma_{k;j} = \tan^{-1} \left( {\frac{y_{k;j}^N - y_{k-1;j}^N}{y_{k;j}^E - y_{k-1;j}^E}} \right).      \end{equation*}     The  information gain on the states of target $i$ at time $k$ as a result of the agents taking actions ${\pmb \xi}^{\rm all}_k$  is then defined as the difference between these measures of information,   \begin{equation} \label{eqn:InfoDifference} \log | {\bf F}_{k|k;i}^{\rm Total} ( {\pmb \xi}_k^{\rm all} ) | - \log | {\bf F}_{k|k-1;i}^{\rm Total} |. \end{equation}  

With regard to these quantities, observe that the post-action Fisher information ${\bf F}_{k|k;i}^{\rm Total}( {\pmb \xi}_k^{\rm all})$ is random. It is random since it depends on knowing which agents will detect target $i$ after all actions are executed at time step $k$.    The post-action Fisher information in turn depends on the unknown positions of the targets and each agent's imperfect detection capability.  Any loss function depending on ${\bf F}_{k|k;i}^{\rm Total} ({\pmb \xi}_k^{\rm all})$ will thus be stochastic.

The loss function each agent hopes to minimize at time $k$ will be denoted as  $L_k \left( {{\pmb \xi}}_k^{\rm all} \right)$. This loss function  is defined as the negative of the total information gain written in (\ref{eqn:InfoDifference}), summed over the total number of targets, $T$.    \begin{equation} L_k({\pmb \xi}_k^{\rm all})= - \sum_{i=1}^T \left( \log \left| {\bf F}_{k|k;i}^{\rm Total} ({\pmb \xi}_k^{\rm all}) \right| - \log \left| {\bf  F}_{k|k-1;i}^{\rm Total} \right| \right).   \label{eqn:TrueLoss}
\end{equation}  The total information gain resulting from the actions  ${\pmb \xi}_k^{\rm all}$ cannot be known in advance of  (i) all of the agents carrying out their respective actions, (ii) all targets moving (randomly) over the time step interval, and (iii)  all agent sensors imperfectly detecting and measuring targets.  The agents must thus minimize $L_k$ while only being able to calculate estimated and stochastic values of the loss function.

To execute the optimization within their respective DECIDE steps, each agent needs to predict the loss function.  Each agent must therefore first predict the post-action Fisher information,  ${\bf F}_{k|k;i}^{\rm Total}({\pmb \xi}_k^{\rm all}).$   It does this by replacing ${\pmb \xi}_k^{\rm all}$ with its predicted values of these components.    It is specifically using its knowledge of the other agents' positions, orientations, and FIMs (communicated to it at time $k-1$) to arrive at a sensible guess of what actions the other agents will take at time $k$.  With this prediction, each agent will have a prediction of the post-action Fisher information.   The post-action Fisher information of target $i$ predicted by agent $j$ as a function of agent $j$'s actions at time $k$  will be denoted as ${\hat {\bf F}}_{k|k;i;j}^{\rm Total}   \left( {\pmb \xi}_{k;j} \right),$ and this will be calculated as $${\hat {\bf F}}_{k|k;i;j}^{\rm Total}   \left( {\pmb \xi}_{k;j} \right) = {\bf F}_{k|k;i}^{\rm Total}  \left.  \left(   {\pmb \xi}_k^{\rm all} \right) \right|_{{\pmb \xi}_{k}^{{\rm all}\backslash j} = {\hat {\pmb \xi}}_{k;j}^{{\rm all}\backslash j}},$$    where ${\pmb \xi}_{k}^{{\rm all} \backslash j} = \left( {\pmb \xi}_{k;1}, \ldots, {\pmb \xi}_{k;j-1}, {\pmb \xi}_{k;j+1}, \ldots, {\pmb \xi}_{k;A} \right)$,   ${\hat {\pmb \xi}}_{k;j}^{{\rm all} \backslash j}  =  \left( { {\pmb \xi}}_{k;1;j}, \ldots, { {\pmb \xi}}_{k;j-1;j}, {\hat {\pmb \xi}}_{k;j+1;j}, \ldots, {\hat {\pmb \xi}}_{k;A;j} \right)$, and   ${\hat {\pmb \xi}}_{k;l;j}$ is the action agent $j$ predicts of agent $l$ at time $k$ .    A more detailed formula for ${\hat {\bf F}}^{\rm Total}_{k|k;i;j}$ is given below in (\ref{eqn:EstimatedFHat}).   \begin{eqnarray}   \nonumber   \lefteqn{ {\hat{\bf  F}}_{k|k;i;j}^{\rm Total}  \left(   {\pmb \xi}_{k;j} \right) } \\  \label{eqn:EstimatedFHat} & = &   {\bf F}_{k|k-1;i;j} +  \\ \nonumber & &     {\hat {\pi}}^d_{k;i;j} \left( {\pmb \xi}_{k;j} \right) ~ \left[ {\hat  {\bf H}}_{k; i;j} \left( {\pmb \xi}_{k;j} \right) \right]^T      {\bf R}^{-1}   {\hat {\bf H}}_{k; i;j} \left(  {\pmb \xi}_{k;j} \right)    +       \\   \nonumber  &  & +  \sum_{l=1; l \neq j}^A   \left \{    {\bf F}_{k|k-1;i;l} {\mathbbm 1} \left( C_{k-1;l;j} = 1 \right)   +  {\hat \pi}^d_{k;i,l;j} \left( {\hat {\pmb \xi}}_{k;l;j} \right)    \right.  \\    \nonumber & &   \times \left.      \left[ {\hat  {\bf H}}_{k; i;l; j} \left( {\hat {\pmb \xi}}_{k;l;j} \right) \right]^T    {\bf R}^{-1}   {\hat {\bf H}}_{k; i;l;j} \left(  {\hat {\pmb \xi}}_{k;l; j} \right)   \right \},  \end{eqnarray} where $C_{k-1;l;j}$ is 1 if agents $l$ and $j$ communicate at time $k-1$ and 0 otherwise, ${\hat \pi}^d_{k;i;l;j} \left( {\hat {\pmb \xi}}_{k;l;j} \right)$ is the probability that agent $l$ detects target $i$, as predicted by agent $j$, and  the matrix  $\hat{\bf{H}}_{k;i;l;j}( {\hat {\pmb{\xi}}}_{k;l;j})$ is the measurement sensitivity Jacobian computed by agent $j$ using its predicted positions of agent $l$ and target $i$.    Note that the first two terms on the right-side of the equality in   (\ref{eqn:EstimatedFHat}) measure agent $j$'s contribution to the FIM of target $i$. The first term is agent $j'$s pre-action Fisher and the second is the additional information resulting from action ${\pmb \xi}_{k;j}$. The last two  terms on the right of (\ref{eqn:EstimatedFHat}) estimate the other agents'  contribution to the Fisher of target $i$ (as predicted by agent $j$). The second-to-last term is agent $l$'s pre-action Fisher on target $i$ and the fourth is the additional information on target $i$ resulting from agent $l$ taking the action agent $j$ would expect it to. We would like to remind the reader, again, that this estimate of the post-action Fisher assumes independence among the measurements.

With the estimated FIM calculated as it is in (\ref{eqn:EstimatedFHat}), agent $j$ calculates a predicted post-action loss value by summing the information losses over all of the targets. This predicted post-action loss of agent $j$'s is denoted as ${\hat L}_{k;j} \left( {\pmb \xi}_{k;j} \right)$ and is calculated as 
\begin{equation}
  \hat{L}_{k;j} \left(   {\pmb \xi}_{k;j}  \right) =  - \sum_{i=1}^T \left( \log \left| \hat{{\bf F}}_{k|k;i;j}^{\rm Total} \left(  {\pmb \xi}_{k;j}  \right)  \right|     -  \log | {\bf F}_{k|k-1;i}^{\rm Total} | \right).  \label{eqn:biased_Lhat}   
\end{equation} Recall this is an approximation to the actual loss.   The value of  $\hat{L}_{k;j}$ can be thought of as $L_k$ conditioned on agent $j$'s predicted positions of all targets and agents.   Hernandez (Ref. \cite{Hernandez}) gives the conditions on Eqn. (\ref{eqn:biased_Lhat}) which guarantee convergence of the CSO when trying to minimize such a function. These conditions do not necessarily apply and can be severely relaxed in our situation, however.  If each agent were iteratively making multiple moves to minimize one stochastically measured loss function, the conditions on Eqn. (\ref{eqn:biased_Lhat}) given in Ref. \cite{Hernandez} would apply. This is not the case, however.  Recall that at each time step, each agent only has the time to make one move (not several moves), and all that is necessary is that this move sufficiently goes in the right direction towards minimizing the loss function.

Before we discuss how the loss function at each time step is minimized, it is important that we mention that ${\hat L}_{k;j}$ is a biased predictor of  $L_k$.    The estimated loss and/or its gradient must be an unbiased predictor of the true (unknown) loss or gradient (see Ref. \cite{Spall_Book}, Chaps. 4--7) for classic stochastic optimization results to hold.   If the probability distributions of $L_k$ and ${\hat L}_{k;j}$ are the same, then $ {\mathbbm E} \left( L_k - {\hat L}_{k;j} \right) = 0$, making ${\hat L}_{k;j}$ unbiased.    We discuss an approximate resolution to this issue below.

 Note that the only distinction between $L_k$ and $\hat{L}_{k;j}$  is the different form for the updated Fisher information. For $L_k$, we use ${\boldsymbol{F}}_{k\vert k;i}^{\rm Total} (\boldsymbol{\xi }_k)$, while for $\hat{L}_{k;j}$, we use ${\hat{\boldsymbol{F}}}_{k\vert k;i;j}^{\rm Total} $.     The probability distributions of $L_k$ and $\hat{L}_k$ will therefore be identical  if the distributions of ${\boldsymbol{F}}_{k\vert k;i}^{\rm Total} (\boldsymbol{\xi }_k)$ and ${\hat{\boldsymbol{F}}}_{k\vert k;i;j}^{\rm Total} ({\hat{\boldsymbol{\xi}}}_k)$ are identical.

 We now describe a method by which we attempt to mitigate the bias issue. We first discuss the method in the idealized case of a linear state-space model with Gaussian randomness. We then offer comments relative to our more practical nonlinear setting. 

From standard orthogonality properties and notation of the generic Kalman filter, it is known that with linear state-space models and Gaussian randomness, the one-step ahead prediction ${\hat{\boldsymbol{x}}}_{k\vert {k-1}}$  is independent of the estimation error $ {\boldsymbol \epsilon}_{k|k-1} = {\boldsymbol{x}}_{k}-{\hat{\boldsymbol{x}}}_{k\vert {k-1}}$.  The standard one-step-ahead error-covariance matrix calculated in the Kalman filter, ${\bf P}_{k|k-1}$, is the covariance of this estimation error, i.e.,  \begin{equation} \label{epsDstn} {\boldsymbol \epsilon}_{k|k-1} \sim N \left(   {\bf 0}, {\bf P}_{k|k-1} \right). \end{equation}    And with the distribution given in (\ref{epsDstn}), one can create a simulated true state by adding a simulated value of ${\boldsymbol{\epsilon}}_k$ to the prediction, ${\hat{\boldsymbol{x}}}_{k\vert {k-1}}$. In the case of the linear state-space model with Gaussian errors,  this simulated true state has a distribution identical to the unknown true state, ${\bf x}_k$.    

In the context of the problem discussed here, we let ${\bf x}_{k|k-1;i;j}$  be agent $j$'s simulated true state  of target $i$  at time step $k$.   That is, each of these will be states formed by adding the above-mentioned Monte Carlo randomness to the corresponding one-step ahead filter estimates. To state this mathematically, \begin{equation}\label{eqn:MC_adjstmnt} {\bf x}_{k|k-1;i;j} = {\hat {\bf x}}_{k|k-1;i;j} + {{\boldsymbol \epsilon}}_{k|k-1;i;j}, \end{equation}  where ${\boldsymbol \epsilon}_{k|k-1;i;j} \sim N \left( {\bf 0}_{6 \times 1}, {\bf P}_{k|k-1;i;j} \right).$

Although the adjustment in (\ref{eqn:MC_adjstmnt}) makes  ${\mathbbm E} \left( L_k- \hat{L}_{k;j} \right)$ = 0 when the state-space model is fully linear, the actual case of interest here involves a nonlinear measurement equation.  The Monte Carlo adjustment to produce a simulated true state is thus an approximation that is valid to the extent that the EKF acts like a standard Kalman filter.

Now that each agent can estimate a  loss function, it can decide (in its DECIDE step) which heading to take and how to vertically displace itself at time $k$.   As mentioned earlier, this involves agent $j$ minimizing ${\hat L}_{k;j} \left( {\pmb \xi}_{k;j} \right)$ with respect to $\gamma_{k;j}$ and $y^U_{k;j}$.   Agent $j$ selects its heading and vertical displacement at time $k$ by taking the stochastic gradient of ${\hat L}_{k;j} $ with respect to $\gamma_{k;j}$ and $y^U_{k;j}$, respectively.     The heading and vertical displacement agent $j$ select at time $k$ are calculated as   \begin{eqnarray} \label{eqn:gammaDeriv} \gamma_{k;j} & = &  \gamma_{k-1;j} - a_k   \left. {\frac{ \partial {\hat L}_{k;j}   \left(  \gamma,~y^U    \right)}{\partial \gamma}} \right|_{\gamma = {\gamma_{k-1;j}}, y^U = y^U_{k-1;j}}, \\ \nonumber {\rm and} & &  \\ \label{eqn:uDeriv}  y^U_{k;j} & = &  y^U _{k-1;j} - b_k   \left. {\frac{ \partial {\hat L}_{k;j}   \left(  \gamma,~y^U    \right)}{\partial y^U}} \right|_{\gamma = {\gamma_{k;j}}, y^U = y^U_{k-1;j}}, \end{eqnarray} where $a_k$ and $b_k$ are  predefined sequences of numbers selected by the user.  This is done iteratively (using the seesaw method), until some convergence threshold or fixed maximum number of iterations is reached.   The details of the seesaw method are given in the next section, and the actual calculations for the derivatives in (\ref{eqn:gammaDeriv}) and (\ref{eqn:uDeriv}) are given in Sections \ref{sctn:Appendix_H} -  \ref{sctn:Appendix_uDeriv} of the Appendix.

\section{Cyclic Stochastic Optimization and the Seesaw Method}
\label{sec:cso-seesaw}


When the number of parameters is large, the problem of jointly minimizing a loss function with respect to  multiple parameters can incur excessively high computational cost.  This computational cost can be reduced, however, using conditional optimization methods. In such methods, the loss function is minimized with respect to a subset of the parameters while the rest of the parameters are held fixed.  Cyclic optimization seeks to combine algorithms for performing conditional optimization with the hopes of obtaining a solution to the joint optimization problem \cite{Spall}. A specific case of cyclic optimization is CSO, which applies when the loss function to be minimized is random or measured with error. 

The CSO technique employed in this project is the seesaw method. In the seesaw method, each agent selects an action in an attempt to minimize its contribution to the loss function. The agent does this given  knowledge (possibly imperfect knowledge) of other agents. Before attempting to minimize the loss function, however, each agent iteratively incorporates the other agents’ information and decisions.  

 A desired feature of any algorithm we use is that each agents' local estimates and control actions converge to the global optimum.    This convergence should occur  under reasonable conditions, and these conditions are usually related to how well the agents communicate.  Two desirable aspects of the seesaw method are: (i) it is a hybrid technique that can operate in either a centralized or decentralized mode, and (ii) the formal structure lends itself to rigorous empirical and theoretical analysis. 

The seesaw method specifically divides the entire parameter (or decision)  vector into at least two subvectors. Each subvector corresponds to the parameters/decisions associated with one of the agents.  As an example, consider a special case where two agents cooperate to minimize a  loss function. In this example we will assume this loss does not depend on time (and will thus not be indexed by $k$).   We will write this loss function as $L$, and $L$ will depend on  the decisions of the two agents,  ${\pmb \beta}^{\rm all}$, where $${\pmb \beta}^{\rm all} = \left( {\pmb \beta}_1^T, {\pmb \beta}_2^T \right)^T,$$ and ${\pmb \beta}_j$ is the decision vector associated with agent $j$.  Mathematically we write the loss function as $$L = L \left( {\pmb \beta}^{\rm all} \right).$$   

 In the seesaw process, iteration by iteration, subvector ${\pmb \beta}_1$ is improved (possibly optimized) conditioned on the most recent value of ${\pmb \beta}_2$ and, likewise, ${\pmb \beta}_2$ is improved (possibly optimized) based on the most recent value of $ {\pmb \beta}_1$. The estimate at the very end of  seesaw iteration $t$ thus has the form 
$$\hat{\pmb \beta}^{{\rm all,~ssaw}_t} \equiv   \left[   \left( \hat{\pmb \beta}_1^{{\rm ssaw}_t} \right)^T ~ \left( \hat{\pmb \beta}_2^{{\rm ssaw}_t} \right)^T    \right]^{T},$$
with ${\hat{\pmb \beta}}_1^{{\rm ssaw}_t}$  a function of  $\hat{\pmb \beta}_1^{{\rm ssaw}_{t-1}}$  and $\hat{\pmb \beta}_2^{{\rm ssaw}_{t-1}}$, and $\hat{\pmb \beta}_2^{{\rm ssaw}_t}$  a function of  $\hat{\pmb \beta}_2^{{\rm ssaw}_{t-1}}$  and $\hat{\pmb \beta}_1^{{\rm ssaw}_t}$.  While this scheme allows for the localized (decentralized) optimization of each subvector, it does not allow for the optimization of the overall decision vector ${\pmb \beta}$. This is not always true when the loss function is not stochastic.    


The above ideas apply directly  to an $M$-agent process, where $M > 2$.  In the case of $M$ agents, suppose that $${\pmb \beta}^{\rm all} = \left[   {\pmb \beta}_1^T,~{\pmb \beta}_2^T,~{\pmb \beta}_3^T,~\cdots~{\pmb \beta}_M^T \right]^T,$$ where ${\pmb \beta}_1, {\pmb \beta}_2, \ldots, {\pmb \beta}_M$ are subevectors, each subvector corresponding to one agent's decision variables, and each processed in the exact same sequential manner as the two-stage algorithm described above.    That is, the vectors are processed sequentially such that 
\begin{equation}
\begin{aligned}
L(\hat{\pmb \beta}^{{\rm all,~ ssaw}_{t+1}}) & \leq \ldots \\
                                     & \leq L(\hat{\pmb \beta}_1^{{\rm ssaw}_{t+1}}, \hat{\pmb \beta}_2^{{\rm ssaw}_{t+1}},  \hat{\pmb \beta}_3^{{\rm ssaw}_t}, \ldots, \hat{\pmb \beta}_M^{{\rm ssaw}_n})                      \\                
                                   &  \leq L(\hat{\pmb \beta}_1^{{\rm ssaw}_{t+1}}, \hat{\pmb \beta}_2^{{\rm ssaw}_t}, \ldots, \hat{\pmb \beta}_M^{{\rm ssaw}_t}) \\
                                   &  \leq L(\hat{\pmb \beta}^{{\rm all,~ssaw}_t})
\end{aligned}
\end{equation}
subject to $\hat{\pmb \beta}^{{\rm all,~ssaw}_{t+1}} \neq \hat{\pmb \beta}^{{\rm ssaw}_t}$ only if $L(\hat{\pmb \beta}^{{\rm all,~ ssaw}_{t+1}}) < L(\hat{\pmb \beta}^{{\rm ssaw}_t})$.

Because of the imperfect information each agent has about the entire system, the seesaw process we employ in this problem has to work with a stochastic loss function.   In particular, each agent will have a “noisy” estimate of the global loss $L$.   This noisy estimate of $L$ can be generically written as ${\hat L} \left( {\pmb \beta}^{\rm all} \right)$, where
\begin{equation}
\hat{L}({\pmb \beta}^{\rm all}) \equiv L({\pmb \beta}^{\rm all}) + \varepsilon({\pmb \beta}^{\rm all})  \label{eqn:GenericLHat}
\end{equation}
and $\varepsilon({\pmb \beta}^{\rm all})$ represents the error due to random quantities such as imperfect state estimates.  At each iteration of the seesaw process, each agent selects an action (a value of ${\pmb \beta}_j$) in an attempt to minimize its contribution to $\hat{L}({\pmb \beta}^{\rm all})$.  The overall aim is to minimize the unknown (ideal) loss $L$ while only using information that is associated with the available $\hat{L}$.    In our surveillance problem, the true loss in (\ref{eqn:TrueLoss}) and the noisy loss in (\ref{eqn:biased_Lhat}) play the role of $L$ and ${\hat L}$, respectively, in (\ref{eqn:GenericLHat}).

At each iteration of the process, the  contribution of each agent may only be improved, versus being optimized.  Ref.~\cite{Spall} shows that convergence to a centralized solution is possible with improvement  at the per-agent level. In fact, convergence is guaranteed if, at each iteration, only one of the $M$ agents improves a sufficient amount.


In the simulations, the agents execute the seesaw within the DECIDE step at time $k$ as follows:   the agents alternately  choose an action by trying to minimize their respective loss functions using their respective current estimates of target states and their predicted actions of peer agents they have yet (or failed) to communicate with.  Each agent predicts each target to move with the same heading and speed as estimated in the previous time step, and each peer agent it has not communicated with is projected to move in the same direction and speed it did in the previous time step.    The seesaw process continues for a prescribed number of iterations, at which point the final decisions are executed.  The process is then repeated for each time step $k$.  

We investigated the performance of our proposed CSO algorithm with simulation studies.    These simulation studies are described in the next section.

\section{Simulation Studies}
\label{sec:simulations}

To test whether the cyclic stochastic optimization algorithm we propose in this paper is effective, we conducted several simulation studies.    In the first  simulation study, we considered two agents and two targets.    In the second and third study, we considered three agents and two targets. In the fourth simulation study, we show how two groups of two agents simultaneously track two targets (note that these two groups of agents are independent of one another, and no communication takes place between the two groups), and in the fifth study, two groups of four agents independently and simultaneously track four targets.    The trajectory of the agents and targets in these simulations are shown in Figures \ref{fig:Simulation1_part1} - \ref{fig:Simulation5_part2}. A small number of agents were selected for these five simulation studies so that the reader can get a clear understanding of how our algorithm performs.  Visually inspecting the trajectories of more than eight agents is difficult to do. In our final studies, however, we do examine the performance of our algorithm when many agents (many more than four) are tracking a set of targets.

In each simulation, the starting location of both the agents and the targets was random (each location was uniformly drawn within an 8 $\times$ 8 $\times$ 8 cube centered at the origin). The motion of the targets throughout each simulation was random and meant to mimic Brownian motion.  At each time point during a simulation, each target randomly selected a heading (uniformly between $0$ and $2 \pi$) and then moved .1 units of distance in that direction. Each target also selected a vertical displacement at each time point. This was randomly drawn from a uniform distribution on [-.15, .15]. This motion of the targets,  however, is a significant departure from their motion as assumed by the agents.\footnote{As mentioned in the introduction, the setting of this problem is deliberately generic.  If the unmanned agents were drones, which have a maximum speed of approximately 65mph $\approx$ 10ft/.1sec = 30ft/.3sec,  the units of distance could be 100 feet and the time units could be one second, or the units of distance could be 300 feet and the time units could be 3 seconds. If the agents were undersea vehicles, which travel at a maximum speed of over 15 knots (approximately 20mph $ \approx $ 10 yards/sec = 20 yards/2sec), the units of distance could be 100 yards and the units of time 10 seconds, or the units of distance could be 200 yards  yards and the units of time could be 20 seconds.} 

 As stated in Section \ref{sec:EKF}, the agents assumed the targets followed the motion model of ${\bf x}_{k;i} = \Phi {\bf x}_{k-1;i} + {\bf w}_k$, where  $\Phi$ is given in (\ref{eqn:PhiMatrix}), $ \Delta t = .1, $ and  ${\bf w}_k \sim N \left( {\bf 0}_{6 \times 1}, {\bf Q} \right)$ with ${\bf Q} = {\rm diag} \left( .03,  .03,  .03, .01, .01, .01  \right).$  The probability of detection, $\pi^d$, decays with the distance between agent and target according to the function given in (\ref{eqn:piEqn})  \begin{eqnarray}   \label{eqn:piEqn} \pi^d_{k;i;j}  & = &   \exp \left \{ -  \left[ \left( \Delta_{k;i;j}^E \right)^2   +   \left( \Delta_{k;i;i}^N  \right)^2    \right. \right.  \nonumber \\ & & ~~~~~~~~~~~+ \left. \left. \left.  \left( \Delta_{k;i;i}^U  \right)^2 \right] \right/100 \right \}. \end{eqnarray}   The probability agent $j$ successfully communicates with agent $l$ at time $k$ is calculated as \begin{eqnarray} \nonumber \rho_{k;j \rightarrow l} &=  & \exp \left \{ -  \left. \left[ \left( y_{k;j}^E - y_{k;l}^E \right)^2   +   \left( y_{k;j}^N - y_{k;l}^N \right)^2    \right. \right. \right. \\ & & \left. \left. \left. ~~~~~~~~~~+ \left( y_{k;j}^U - y_{k;l}^U \right)^2 \right] \right/200 \right \}.   \label{eqn:communic_Prob}  \end{eqnarray} We select this communication probability to illustrate that our methodology is robust to non-ideal communications. The communications are non-ideal since, at a specific distance, the probability of two agents successfully communicating is only slightly larger than the probability of an agent detecting an adversarial vehicle. This is shown in Figure \ref{fig:ProbDtctAndCmnct}. In Section \ref{sctn:CmnctProb} of the Appendix, we illustrate how the algorithm would perform with more reliable communications.

{
\begin{figure}[h]
\begin{center}
{\includegraphics[trim = 1cm 7cm 1cm 2cm, scale = 0.45]{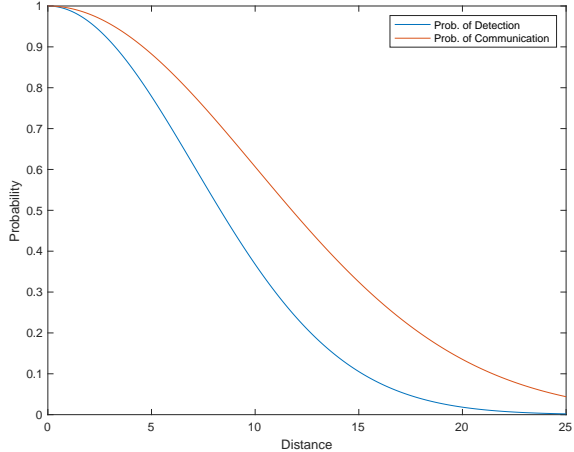}}

\caption{The probability of detection, $\pi^d_{k;i;j},$ and the probability of communication, $\rho_{k;j\rightarrow l}(200),$ as a function of distance.}

\label{fig:ProbDtctAndCmnct}
\end{center}
\end{figure}
}

The values of $\sigma_{\theta}$, $\sigma_{\phi}$, and $\sigma_r$ are all .01,  and the step-sizes in the stochastic gradient calculation, $a_k$ and $b_k$, were set to 1  and 0.1, respectively, for all values of $k$.      Images of the first five simulation studies are given in Figures \ref{fig:Simulation1_part1} - \ref{fig:Simulation5_part2}.     The paths of both targets are given in different shades of black (in the North-East plots their starting positions are circles and their final positions are squares), and the trajectories of the agents are given in red, blue, and green.    Just as with the agents, in the North-East plots their starting positions are circles and their final positions are squares.
   
%

{
\begin{figure}[h]
\begin{center}
{\includegraphics[trim = 1cm 7cm 1cm 2cm, scale = 0.45]{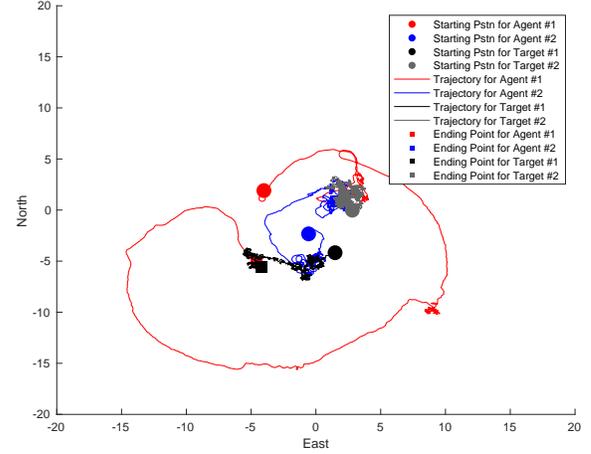}}

\caption{The North-East motion of two targets and two agents across time (1500 time points).  }

\label{fig:Simulation1_part1}
\end{center}
\end{figure}
}

{
\begin{figure}[h]
\begin{center}
{\includegraphics[trim = 1cm 7cm 1cm 7cm, scale = 0.45]{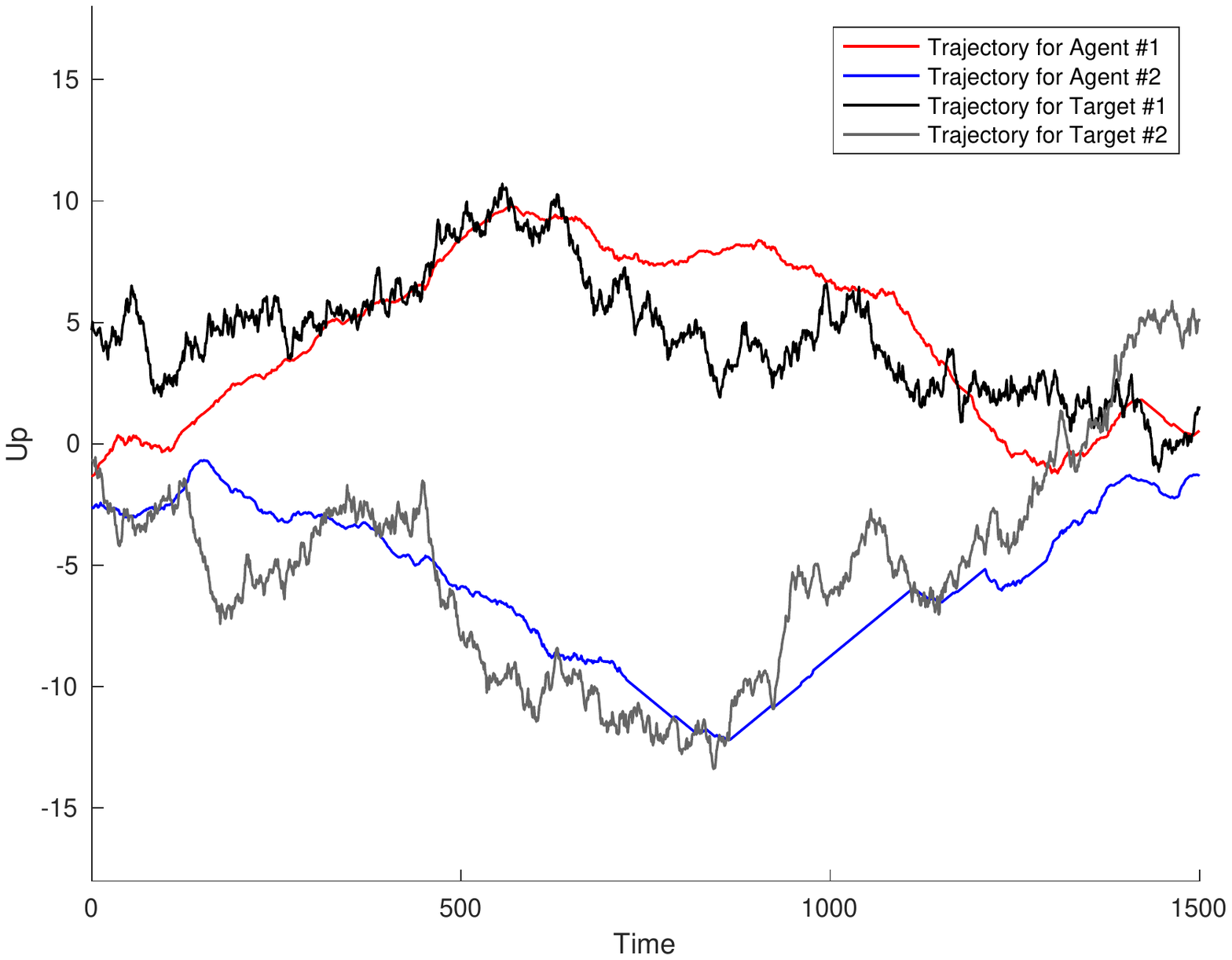}}

\caption{The vertical motion of two targets and two agents across time (1500 time points).}

\label{fig:Simulation1_part2}
\end{center}
\end{figure}
}

{
\begin{figure}[h]
\begin{center}
{\includegraphics[trim = 1cm 7cm 1cm 2cm, scale = 0.45]{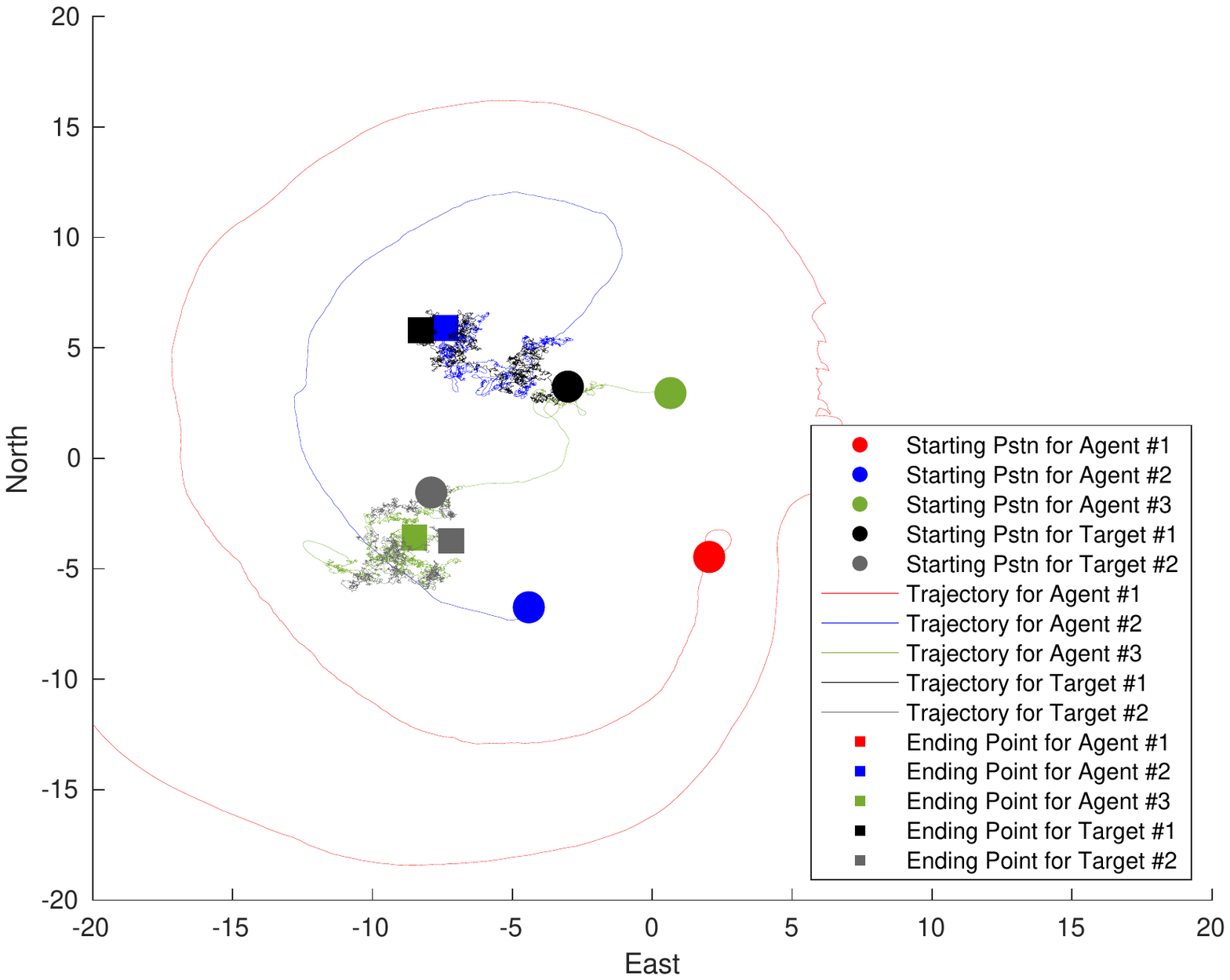}}

\caption{The North-East motion of two targets and three agents across time (2000 time points).  }

\label{fig:Simulation2_part1}
\end{center}
\end{figure}
}

{
\begin{figure}[h]
\begin{center}
{\includegraphics[trim = 1cm 7cm 1cm 7cm, scale = 0.45]{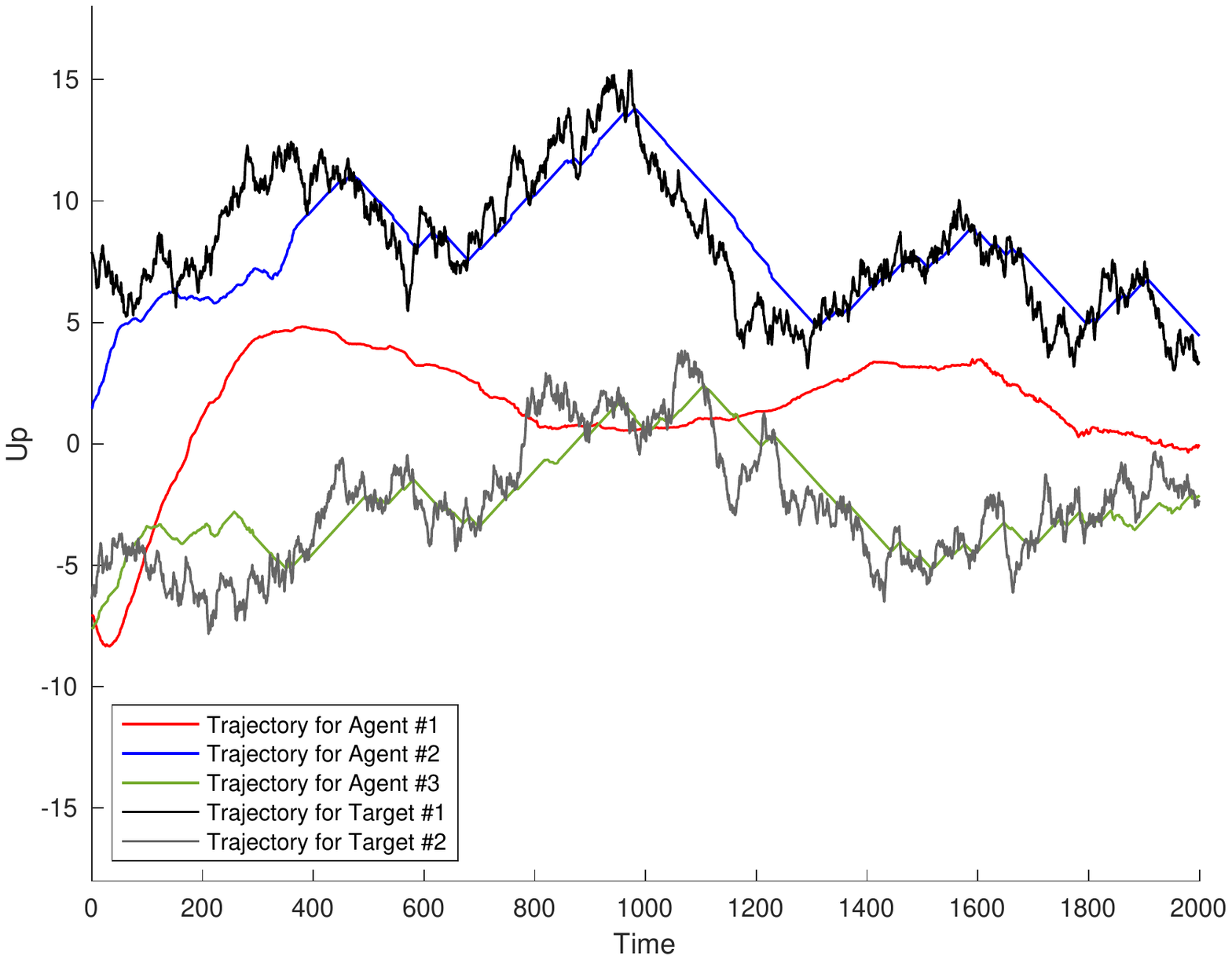}}

\caption{The vertical motion of two targets and three agents across time (2000 time points).}

\label{fig:Simulation2_part2}
\end{center}
\end{figure}
}

{
\begin{figure}[h]
\begin{center}
{\includegraphics[trim = 1cm 7cm 1cm 2cm, scale = 0.45]{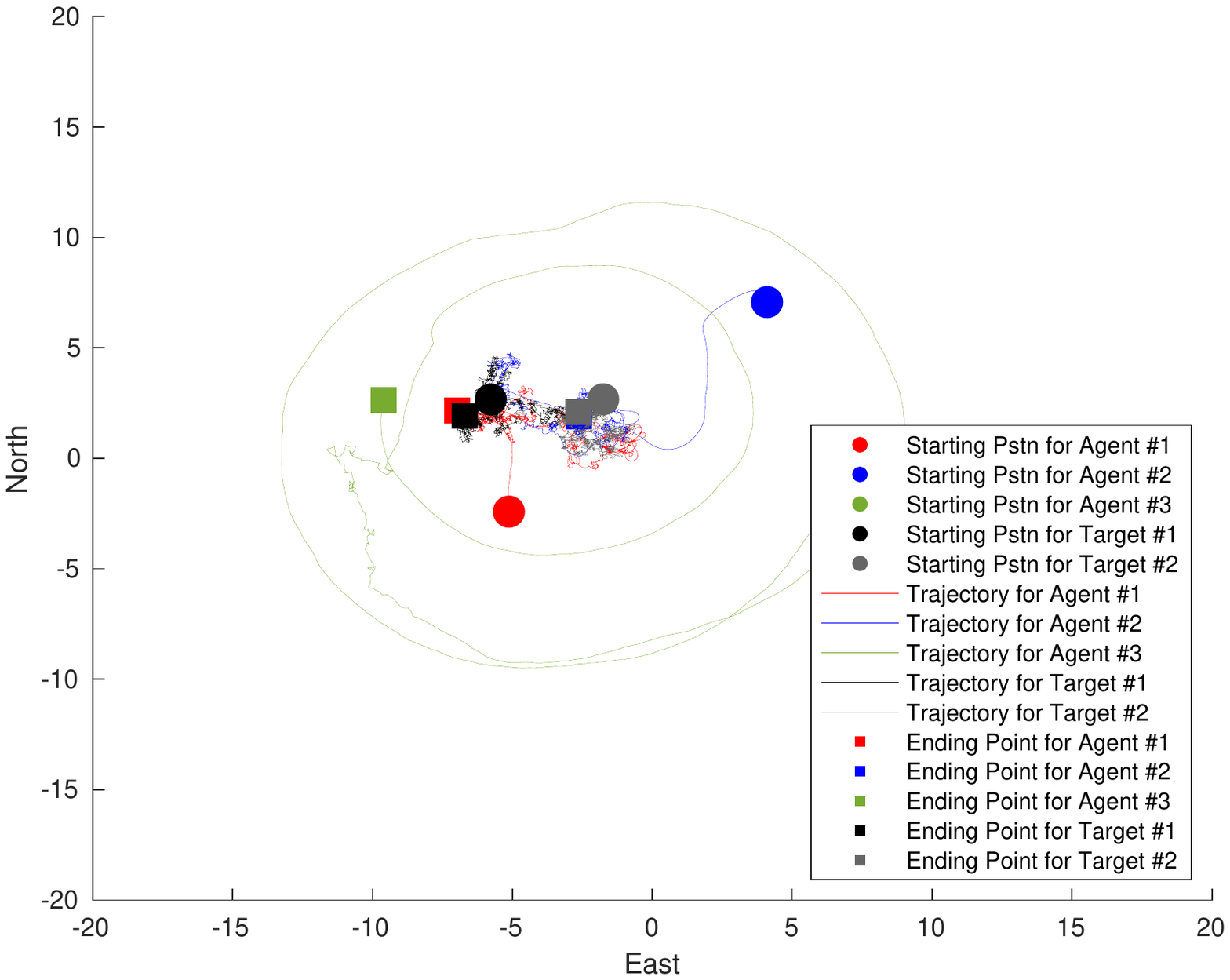}}

\caption{The North-East motion of two targets and three agents across time (1500 time points).  }

\label{fig:Simulation3_part1}
\end{center}
\end{figure}
}

{
\begin{figure}[h]
\begin{center}
{\includegraphics[trim = 1cm 7cm 1cm 7cm, scale = 0.45]{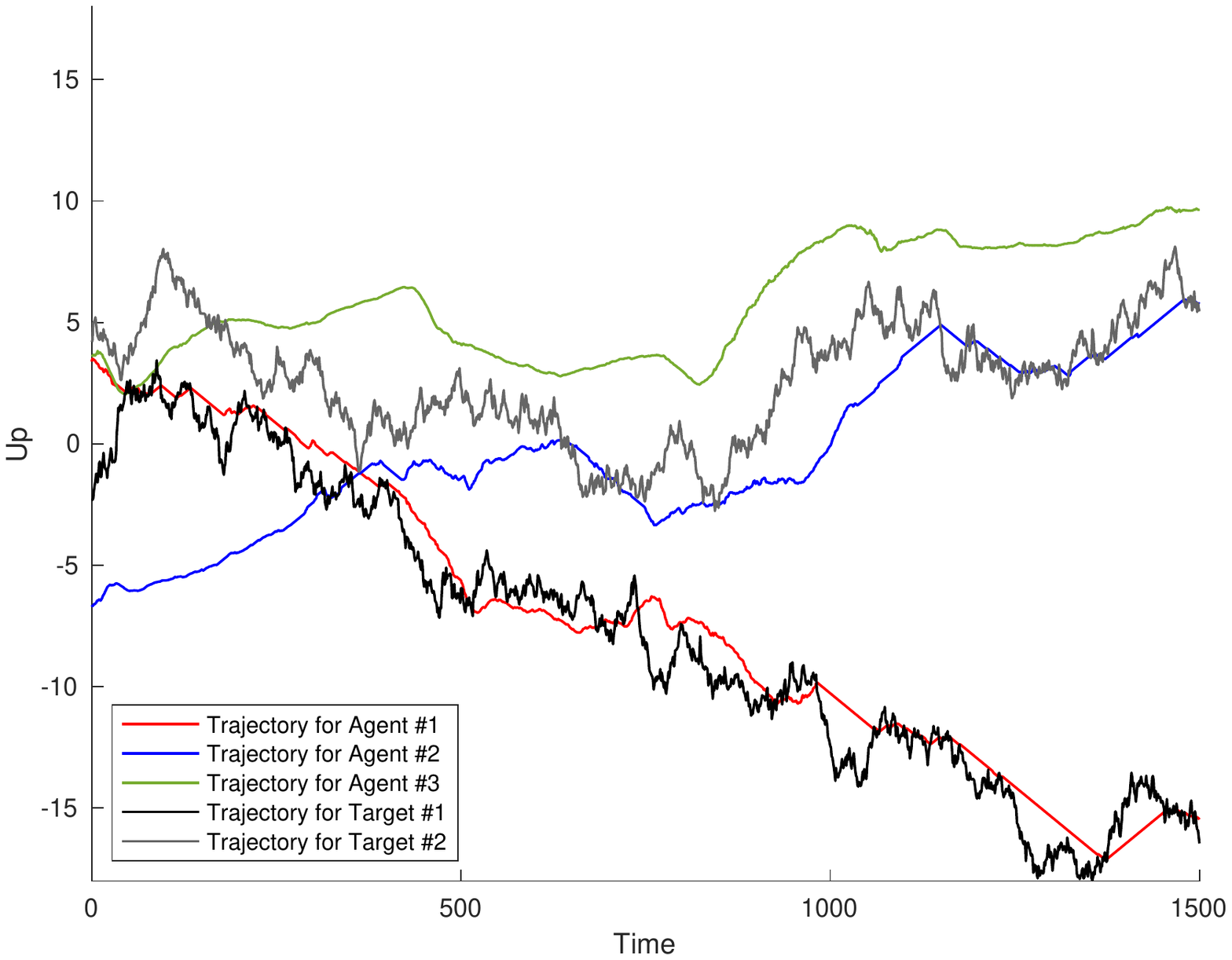}}

\caption{The vertical motion of two targets and three agents across time (1500 time points).}

\label{fig:Simulation3_part2}
\end{center}
\end{figure}
}

{
\begin{figure}[h]
\begin{center}
{\includegraphics[trim = 1cm 7cm 1cm 2cm, scale = 0.45]{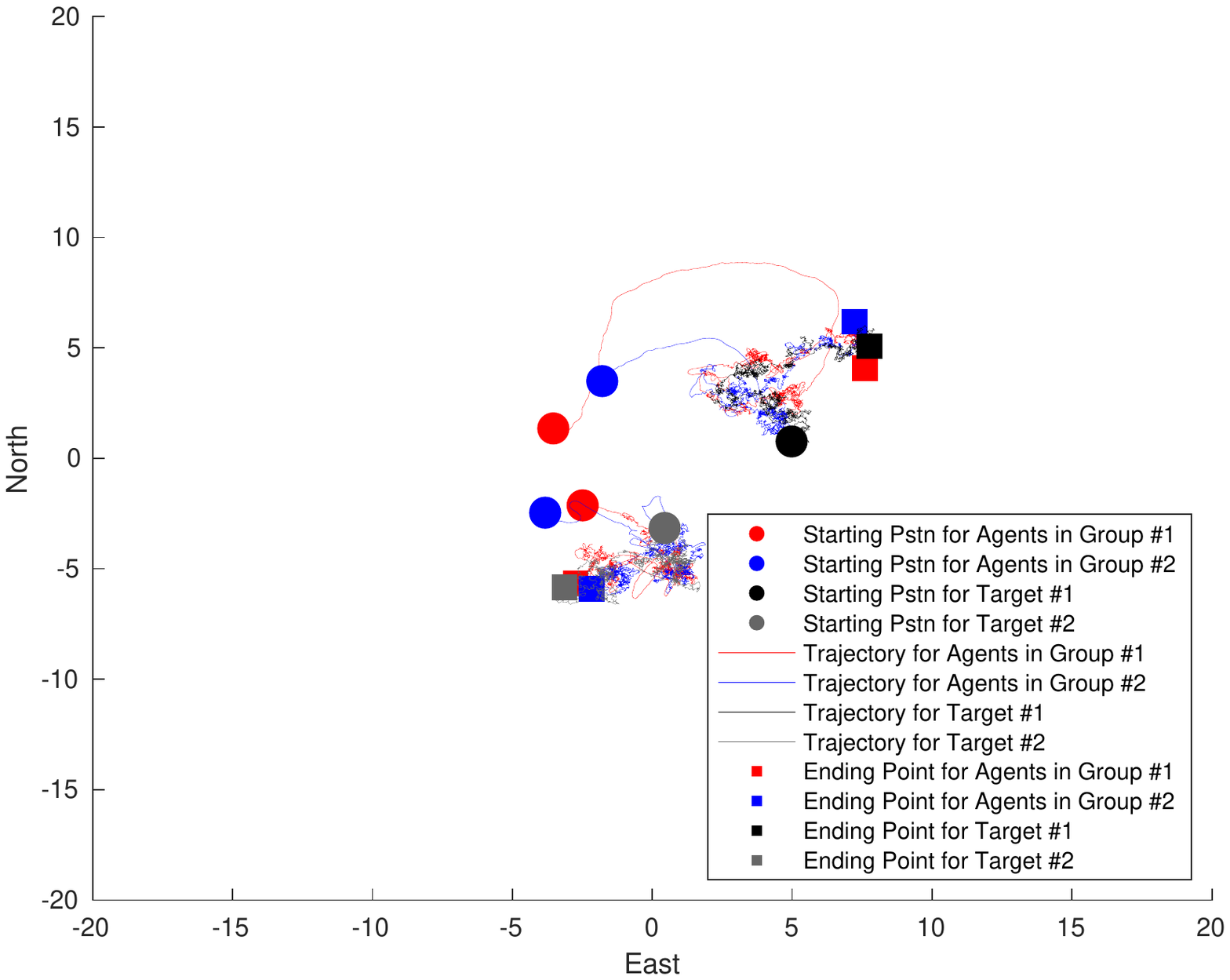}}

\caption{The North-East motion of two targets and two groups of two agents across time (1500 time points).  }

\label{fig:Simulation4_part1}
\end{center}
\end{figure}
}

{
\begin{figure}[h]
\begin{center}
{\includegraphics[trim = 1cm 7cm 1cm 7cm, scale = 0.45]{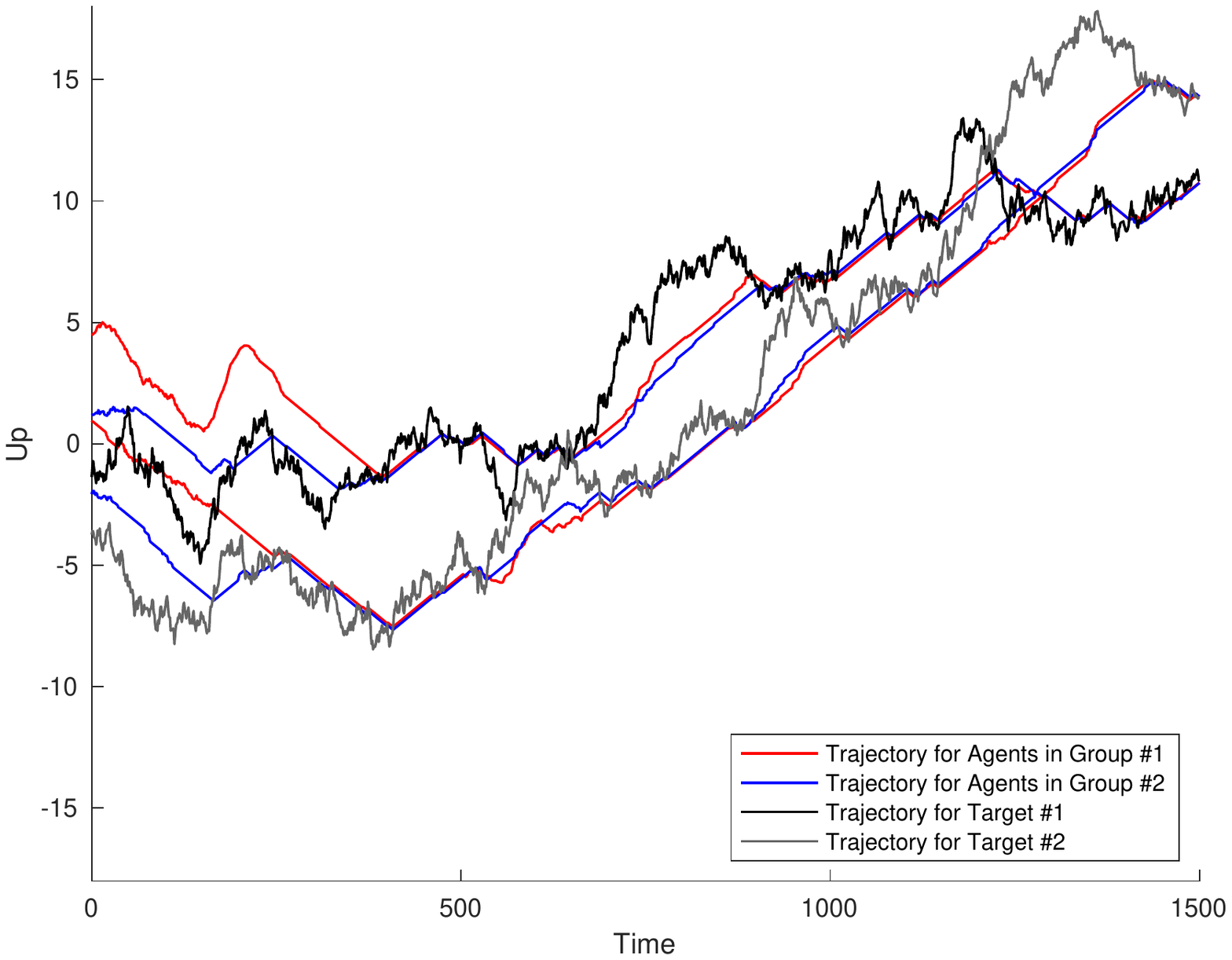}}

\caption{The vertical motion of two targets and two groups of two agents across time (1500 time points).}

\label{fig:Simulation4_part2}
\end{center}
\end{figure}
}

{
\begin{figure}[h]
\begin{center}
{\includegraphics[trim = 1cm 7cm 1cm 2cm, scale = 0.45]{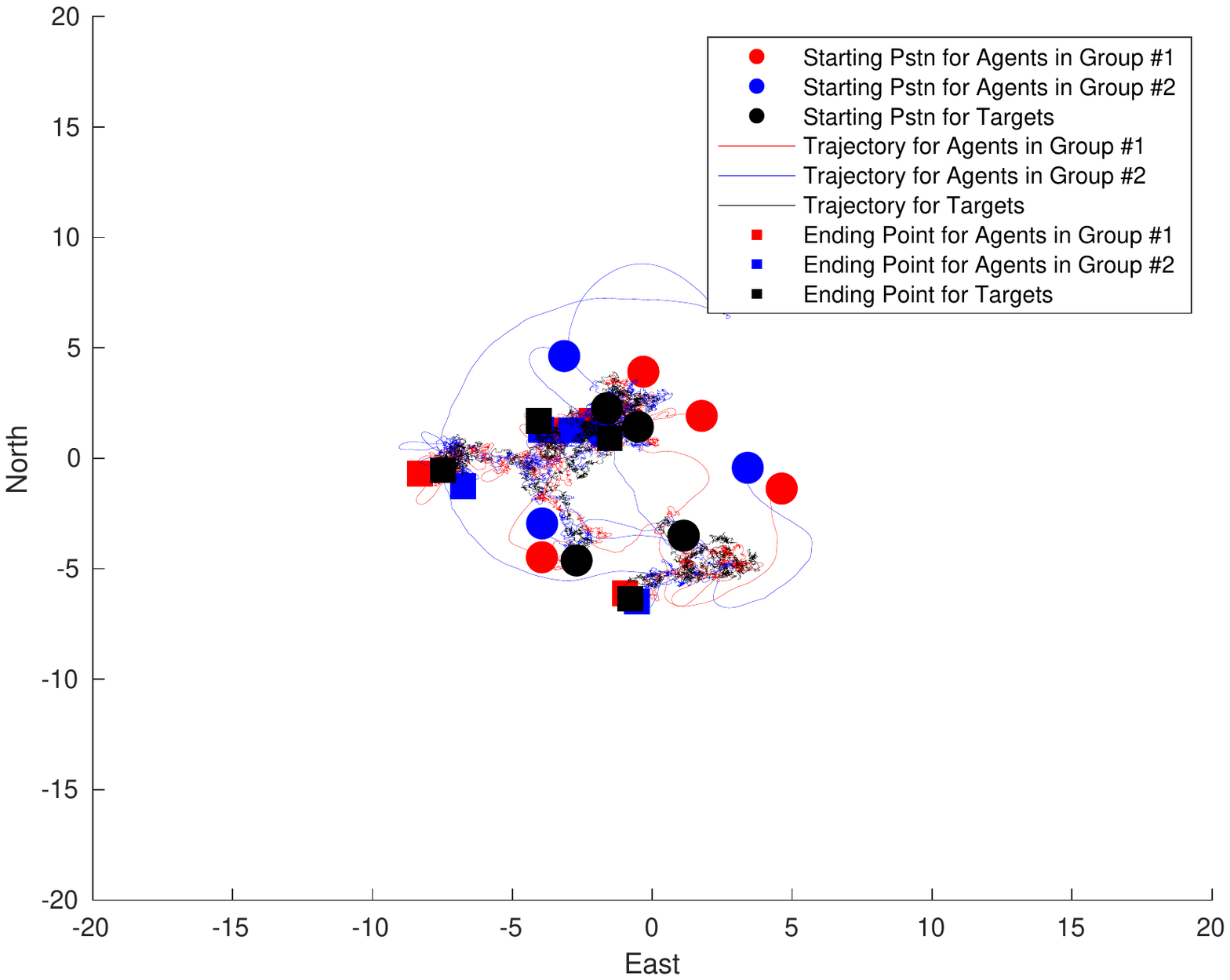}}

\caption{The North-East motion of four targets and two groups of four agents across time (1500 time points).  }

\label{fig:Simulation5_part1}
\end{center}
\end{figure}
}

{
\begin{figure}[h]
\begin{center}
{\includegraphics[trim = 1cm 7cm 1cm 7cm, scale = 0.45]{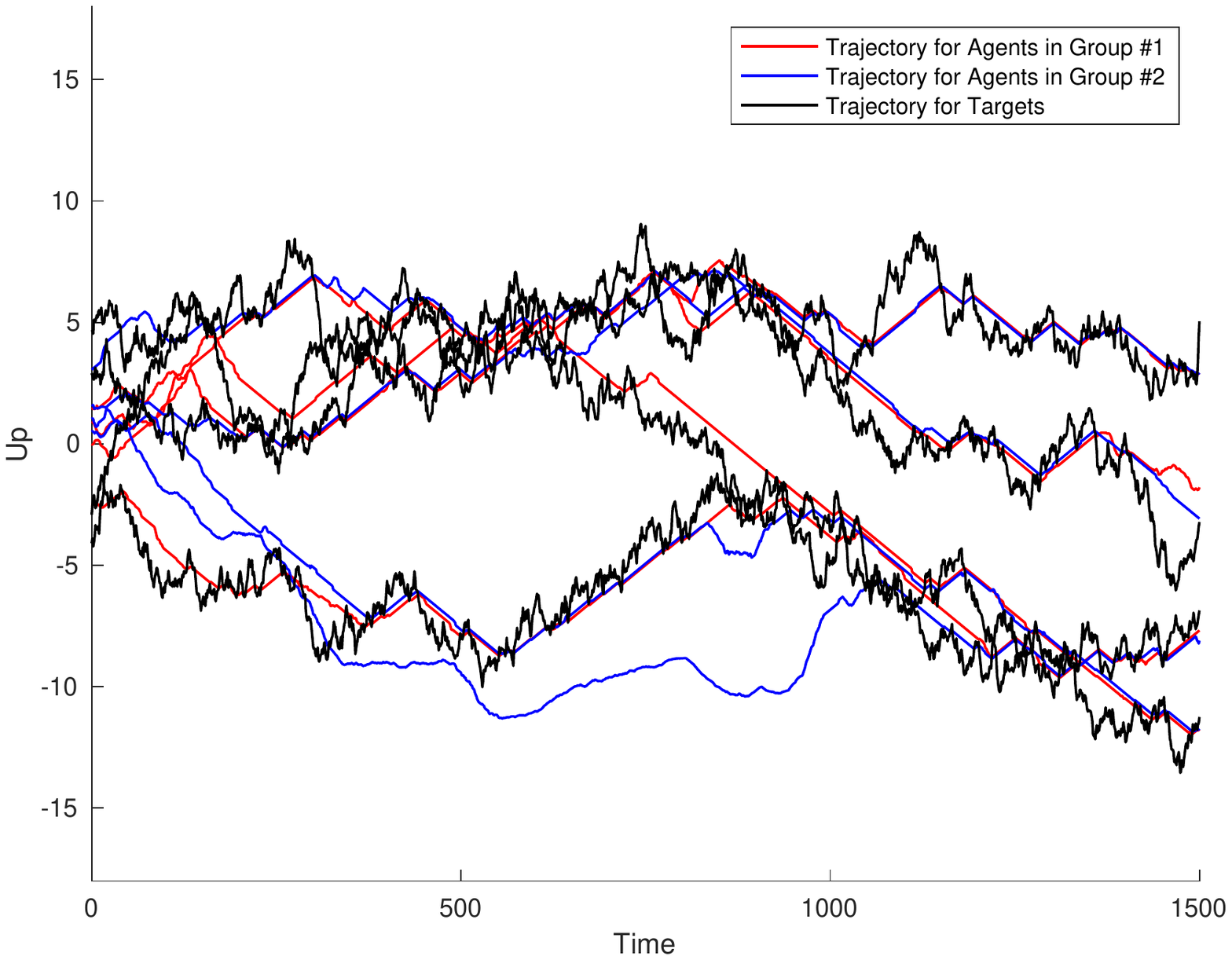}}

\caption{The vertical motion of four targets and two groups of four agents across time (1500 time points).}

\label{fig:Simulation5_part2}
\end{center}
\end{figure}
}

Figures \ref{fig:Simulation1_part1} - \ref{fig:Simulation1_part2} show the results of the first simulation. In these figures  it is clear that the agents effectively track and go in the direction of the targets.  The red agent loops around the other vehicles to eventually track the black target which is moving west. The blue agent travels northeast to track the grey target.   These agents seem to stick with the targets they are following vertically as time progresses as well.  In Figure \ref{fig:Simulation2_part1} - \ref{fig:Simulation2_part2} we have three agents and two targets. Observe how the green agent starts at roughly the same vertical position as the grey target and travels southwest to track it, while the blue agent starts at roughly the same vertical position as the black target and travels northeast to track it.   Also observe that the red agent departs and circles the scene; since both targets have been tracked by the other agents, any additional information the red agent will gain by moving closer  to the any of the two targets is negligible. The red agent thus keeps its distance. This same story is told in Figures \ref{fig:Simulation3_part1} and \ref{fig:Simulation3_part2}; in this case the green agent departs and circles the scene.   

  In Figures \ref{fig:Simulation4_part1} - \ref{fig:Simulation4_part2}, we see what happens when two groups of two agents independently and simultaneously track two targets. One group of agents is in red and the other is in blue.  In this case, the phenomenon observed in Figures \ref{fig:Simulation2_part1} - \ref{fig:Simulation2_part2}  and Figures \ref{fig:Simulation3_part1} - \ref{fig:Simulation3_part2} (one agent departing and circling around the other vehicles) does not occur. This is because the agents that are following the same target are operating independently and not communicating with one another.    Each target is thus very closely followed by two agents.    The same story is told in Figures \ref{fig:Simulation5_part1} - \ref{fig:Simulation5_part2} which illustrate our fifth simulation study. In this case, two groups of four agents (a total of eight agents) independently and simultaneously track four targets.

\begin{figure}[th]
   \begin{center}
   \begin{tabular}{c}
      \includegraphics[trim = 1cm 7cm 1cm 7cm, scale = 0.45]{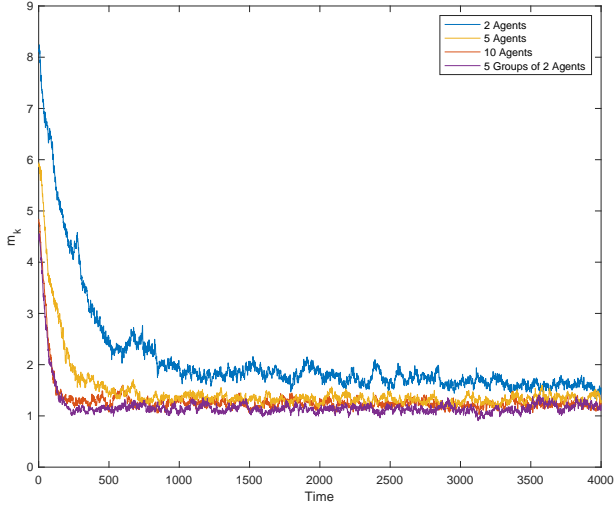}
   \end{tabular}
   \end{center}
\caption{Median distance (across 100 simulations)  from target to closest agent over time. Total number of targets is 2.}
   \label{fig:medianDist_T2} 
\end{figure}

\begin{figure}[th]
   \begin{center}
   \begin{tabular}{c}
      \includegraphics[trim = 1cm 7cm 1cm 7cm, scale = 0.45]{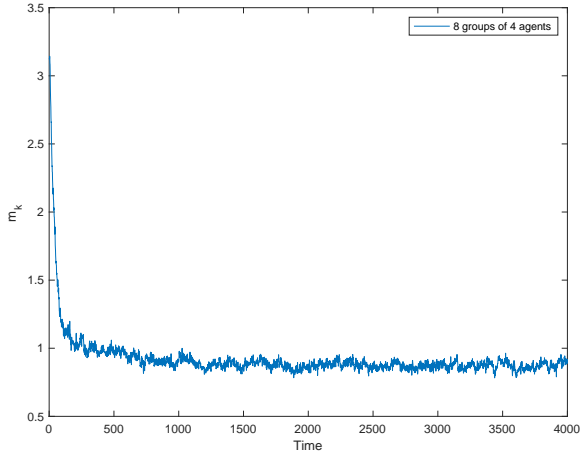}
   \end{tabular}
   \end{center}
\caption{Median distance (across 100 simulations)  from target to closest agent over time. Total number of targets is 4, and the total number of agents is 32 (8 groups of 4 agents).}
   \label{fig:medianDist_T4} 
\end{figure}

Another set of simulation studies were done to uncover the average behavior of multiple agents against a set of targets.    One hundred simulations were done in each study, and the median minimum distance to an agent from each target was calculated at each time point.  We denote this distance at time $k$ as $m_k$.  The results in Figure \ref{fig:medianDist_T2} are when $T$, the total number of targets, is 2, and the result in Figure \ref{fig:medianDist_T4} is when $T = 4$.    Figure \ref{fig:medianDist_T2} shows that when $A = 2$, i.e., when there are two agents communicating with each other, the agents converge towards the targets. This convergence is more pronounced when there are 5 agents tracking two targets. An additional five agents (for a total of 10) only provides a slight improvement, and this is mostly likely a consequence of the behavior observed in Figures \ref{fig:Simulation2_part1} - \ref{fig:Simulation3_part2}. Recall that these figures illustrate that an excessive number of communicating agents does not help in tracking and following targets, as the additional information an extra agent provides on the kinetic states of a target is negligible.  

The simulated setting which provides the fastest and closest tracking is when five groups of two agents independently and simultaneously track the targets. This is a scaled-up version of the simulation shown in Figures \ref{fig:Simulation4_part1} - \ref{fig:Simulation4_part2}, where two groups (as opposed to five)  of two agents independently and simultaneously track two targets. The convergence in this case is shown by the magenta line in Figure \ref{fig:medianDist_T2} and is so prompt and close since (within each group of agents) there are no extraneous agents whose additional information is negligible. This type of convergence is observed on a larger scale in Figure \ref{fig:medianDist_T4}, where eight groups of four agents (a total of 32 agents) track four targets.  This simulation is a scaled-up version of the simulation shown in Figures \ref{fig:Simulation5_part1} - \ref{fig:Simulation5_part2}, where only two (as opposed to eight) groups of four agents independently and simultaneously track four targets.

As can be seen in Figures \ref{fig:Simulation1_part1} - \ref{fig:medianDist_T4}, it appears that, on average, the targets are detected and found relatively quickly.    This implies that the agents consistently choose (via the stochastic gradient method) to be close to a target.     Recall that the closer an agent is to a target, the more likely it is to gain information about the target's states, and information gain on the states of all of the targets is the ultimate goal.

We also explored how the computational cost of the algorithm scales with the number of agents, $A$, and the number of targets, $T$. Table \ref{tab:CPU_time} gives the average CPU time it takes to run one simulation for 4000 time steps (using {\tt MATLAB 2020b} on a MacBook Pro with a 2.3 GHz 8-Core Intel Processor). Table \ref{tab:CPU_time} also gives the average CPU time it takes for one agent to do the necessary filtering and processing in one time step. We refer to the former metric as the simulation time (ST) and the latter metric as Agent Processing Time (APT). The APT may be useful in a practical setting when scheduling the time delay between two different agents executing the PAC steps.

\begin{table}[ht]
\centering

\caption{The average simulation time (ST) and the average agent processing time (APT) in seconds.}

\begin{tabular}{ccccc}

\hline 

A & T &   Mean ST  & Mean APT \\ 

\hline

2 & 2 & 47.86  & .00487  \\

5 & 2 & 203.33  & .00934 \\

5 & 3 & 273.62  & .0127 \\

5 & 4 & 377.56  & .0176 \\

5 & 5 & 498.42  & .0232 \\

5 & 10 & 879.46  & .0412 \\

5 & 15 & 1256.10 & .0584 \\

10 & 2 & 679.91  & .0161 \\

15 & 2 & 1431.37  & .0228 \\

20 & 2 & 2537.79 & .0305 \\

25 & 2 & 3817.83  & .0368 \\
\hline

\end{tabular}

\label{tab:CPU_time}

\end{table}

Plotting mean ST as $T$ increases with the value of $A$ fixed (at $A = 5$) and plotting mean ST as $A$ increases with the value of $T$ fixed (at $T = 2$), it is clear that the computational cost of the simulation grows quadratically with the number of agents and linearly with the number of targets. Plotting mean APT as $T$ increases with the value of $A$ fixed (at $A = 5$) and plotting mean APT as $A$ increases with the value of $T$ fixed (at $T = 2$), it is clear that the computational cost of the agent processing time grows linearly with the number of agents and the number of targets.

\section{Conclusions}
\label{sec:conclusions}  


We have observed in our simulation study that our proposed CSO is effective.   On average, agents do successfully swarm to locate, track, and follow the moving enemy assets.    Our proposed CSO algorithm is also simple.  It requires that each agent iteratively (one-at-a-time) sense the targets, communicate its position and FIMs, estimate a loss function which decreases with decreasing uncertainty in the estimated states, and then selects a motion that minimizes its contribution to this loss function.       Areas of future work include accounting for the dependence in the measurements collected by the agents, and possibly avoiding agent collisions.

\appendix

\allowdisplaybreaks

\subsection{Calculation of ${\bf H}_{k;i;j}$}{\label{sctn:Appendix_H}}

We will let ${\bf H}_{k;i;j}[a,b]$ denote the element of ${\bf H}_{k;i;j}$ in row $a$ and column $b$.   Unless otherwise indicated, ${\bf H}_{k;i;j}[a,b] = 0.$ \begin{eqnarray*} {\bf H}_{k;i;j} \left[1,1 \right] & = &    \left.    {\hat \Delta}_{k|k-1;i;j}^E \right/  {\hat r}_{k|k-1;i;j}  \\  {\bf H}_{k;i;j} \left[1,2 \right] & = &    \left.    {\hat \Delta}_{k|k-1;i;j}^N \right/  {\hat r}_{k|k-1;i;j}  \\  {\bf H}_{k;i;j} \left[1,3 \right] & = &    \left.    {\hat \Delta}_{k|k-1;i;j}^U \right/  {\hat r}_{k|k-1;i;j} \\   {\bf H}_{k;i;j}[2,1] & = & -  \left.  {\hat \Delta}_{k|k-1;i;j}^N  \right/    {\hat f}_{k|k-1;i;j}     \\   {\bf H}_{k;i;j}[2,2] & = &   \left.    {\hat \Delta}_{k|k-1;i;j}^E   \right/  {\hat f}_{k|k-1;i;j}   \\   {\bf H}_{k;i;j}[3,1] & = &      \left[ 1 -  \left(  \left.    {\hat \Delta}_{k|k-1;i;j}^U \right/  {\hat r}_{k|k-1;i;j} \right)^2 \right]^{-{\frac{1}{2}}}   \times \\ & &   \left. {\hat \Delta}_{k|k-1;i;j}^U   {\hat \Delta}_{k|k-1;i;j}^E   \right/   \left(   {\hat r}_{k|k-1;i;j}  \right)^3 \\   {\bf H}_{k;i;j}[3,2] & = &    \left[ 1 -  \left(  \left.    {\hat \Delta}_{k|k-1;i;j}^U \right/  {\hat r}_{k|k-1;i;j} \right)^2 \right]^{-{\frac{1}{2}}}  \times \\ & &   \left. {\hat \Delta}_{k|k-1;i;j}^U   {\hat \Delta}_{k|k-1;i;j}^N   \right/   \left(   {\hat r}_{k|k-1;i;j}  \right)^3     \\   {\bf H}_{k;i;j}[3,3] & = &  -   \left[ 1 -  \left(  \left.    {\hat \Delta}_{k|k-1;i;j}^U \right/  {\hat r}_{k|k-1;i;j} \right)^2 \right]^{-{\frac{1}{2}}} \times \\  & &    \left[  {\frac{1}{   {\hat r}_{k|k-1;i;j}}} -   {\frac{ \left( {\hat \Delta}_{k|k-1;i;j}^U \right)^2 }  {   \left(  {\hat r}_{k|k-1;i;j}  \right)^3}} \right]     \\ {\hat r}_{k|k-1;i;j}  &  = &      \left[ \left(   {\hat \Delta}_{k|k-1;i;j}^E  \right)^2 + \left(   {\hat \Delta}_{k|k-1;i;j}^N  \right)^2+ \right. \\ & & \left.  \left(   {\hat \Delta}_{k|k-1;i;j}^U \right)^2 \right]^{{\frac{1}{2}}}    \\ {\hat f}_{k|k-1;i;j}  &  = &      \left[ \left(   {\hat \Delta}_{k|k-1;i;j}^E  \right)^2 + \left(   {\hat \Delta}_{k|k-1;i;j}^N  \right)^2 \right]^{\frac{1}{2}}   \\   {\hat \Delta}_{k|k-1;i;j}^E &  = &    {\hat x}_{k|k-1;i;j}^E - y_{k;j}^E   \end{eqnarray*}

\subsection{Equation (\ref{eqn:gammaDeriv})}{\label{sctn:Appendix_gammaDeriv}}

\allowdisplaybreaks
	 \begin{eqnarray*} \lefteqn{   \left.   {\frac{\partial {\hat L}_{k;j}(\gamma, y^U)}{\partial \gamma}} \right|_{\gamma = \gamma_{k-1;j}, y^U = y^U_{k-1;j} }} \\   & = & \left. - \sum_{i=1}^T \left \{     \left| {\hat {\bf F}}_{k|k;i;j}^{\rm Total} \left( {\pmb \xi}_{k;j} \right) \right| \cdot   \right.  \right.  \\ & &  \left. \left.   {\rm Tr} \left[    {\hat {\bf F}}_{k|k;i;j}^{\rm Total} \left( {\pmb \xi}_{k;j} \right) ^{-1} {\frac{\partial  {\hat {\bf F}}_{k|k;i;j}^{\rm Total} \left( {\pmb \xi}_{k;j} \right) }{\partial \gamma}}    \right]    \right \}  \right|_{\gamma = \gamma_{k-1;j}, y^U = y^U_{k-1;j} } \end{eqnarray*} where \begin{eqnarray*}  \lefteqn{   \left.  {\frac{\partial {\hat {\bf F}}_{k|k;i;j} \left( {\pmb \xi}_{k;j} \right)}{\partial \gamma}}     \right|_{\gamma = \gamma_{k-1;j}, y^U = y^U_{k-1;j} } } \\ &     = &     \left[   {\frac{\partial {\hat {\pi }}_{k;i;j}^d \left( {\pmb \xi}_{k;j} \right)}{\partial \gamma}}   \right.    {\hat  {\bf H}}_{k; i;j} \left( {\pmb \xi}_{k;j} \right)^T      {\bf R}^{-1}   {\hat {\bf H}}_{k; i;j} \left(  {\pmb \xi}_{k;j} \right) +   \\  &    &   {\hat \pi}^d_{k;i;j} \left( {\pmb \xi}_{k;j} \right)   \left \{ {\frac{\partial {\hat {\bf H }}_{k;i;j} \left( {\pmb \xi}_{k;j} \right)}{\partial \gamma}}^T     {\bf R}^{-1}    {\hat {\bf H}}_{k; i;j} \left(  {\pmb \xi}_{k;j} \right) +   \right.   \\ & &  \left. \left. \left.        {\hat {\bf H}}_{k; i;j} \left(  {\pmb \xi}_{k;j} \right)^T {\bf R}^{-1}  {\frac{\partial {\hat {\bf H }}_{k;i;j} \left( {\pmb \xi}_{k;j} \right)}{\partial \gamma}}  \right \} \right]  \right|_{\gamma = \gamma_{k-1;j}, y^U = y^U_{k-1;j} },\\  \lefteqn{ \left.  {\frac{\partial {\hat {\pi }}_{k;i;j}^d \left( {\pmb \xi}_{k;j} \right)}{\partial \gamma}} \right|_{\gamma = \gamma_{k-1;j}}} \\  & = & -{\hat {\pi }}_{k;i;j}^d \left( {\pmb \xi}_{k;j} \right) \cdot {\frac{ {\hat r}_{k|k-1;i;j}}{50}} \left( {\frac{\partial {\hat r}_{k|k-1;i;j}}{\partial \gamma_{k-1;j}}} \right),~{\rm  and} \\     \lefteqn{     {\frac{\partial  {\hat r}_{k|k-1;i;j} }{\partial \gamma_{k-1;j}}}  =  \left.   {\frac{\partial r_{k|k-1;i;j}}{\partial \gamma}} \right|_{\gamma= \gamma_{k-1;j}}} \\ &  = &  \left(  {\hat r}_{k|k-1;i;j} \right)^{-1} \left(  {\hat \Delta}_{k|k-1;i;j}^E \sin \left(  \gamma_{k-1;j}  \right)   \right.  \\ & &   \left. -   {\hat \Delta}_{k|k-1;i;j}^N \cos \left(  \gamma_{k-1;j}  \right)  \right)   .\end{eqnarray*}     Below are the elements of   $ \left. {\frac{\partial {\hat {\bf H }}_{k;i;j} \left( {\pmb \xi}_{k;j} \right)}{\partial \gamma}}  \right|_{\gamma = \gamma_{k-1;j}, y^U = y^U_{k-1;j} }.$ For notational simplicity, we denote the $(a,b)^{\rm th}$ element in this matrix as  ${\hat {\bf H}}^{{\rm deriv},~\gamma}_{k;i;j} [a,b],$ and unless otherwise stated, ${\hat {\bf H}}^{{\rm deriv},~\gamma}_{k;i;j} [a,b] = 0.$      \begin{eqnarray*}   \lefteqn{  {\hat {\bf H}}^{{\rm deriv},~\gamma}_{k;i;j} [1,1]  =  -{\frac{ {\hat x}_{k|k-1;i;j}^E} {{\hat r}_{k|k-1;i;j}^2}} \left(  {\frac{\partial {\hat r}_{k|k-1;i;j}}{\partial \gamma_{k-1;j}}} \right)}  \\ & &  - \left[ {\frac{\partial y_{k-1;j}^E}{\partial \gamma_{k-1;j}}} \left( {\frac{1}{{\hat r}_{k|k-1;i;j}}} \right) - {\frac{y_{k-1;j}^E}{ {\hat r}_{k|k-1;i;j}^2}}  \left( {\frac{\partial  {\hat r}_{k|k-1;i;j} }{\partial \gamma_{k-1;j}}} \right) \right] \\     \lefteqn{  {\hat {\bf H}}^{{\rm deriv},~\gamma}_{k;i;j} [1,2]  =  -{\frac{ {\hat x}_{k|k-1;i;j}^N} {{\hat r}_{k|k-1;i;j}^2}} \left(  {\frac{\partial {\hat r}_{k|k-1;i;j}}{\partial \gamma_{k-1;j}}} \right)} \\ & &  - \left[ {\frac{\partial y_{k-1;j}^N}{\partial \gamma_{k-1;j}}} \left( {\frac{1}{{\hat r}_{k|k-1;i;j}}} \right) - {\frac{y_{k-1;j}^N}{ {\hat r}_{k|k-1;i;j}^2}} \left(  {\frac{\partial  {\hat r}_{k|k-1;i;j} }{\partial \gamma_{k-1;j}}} \right) \right]    \\  \lefteqn{     {\hat {\bf H}}^{{\rm deriv},~\gamma}_{k;i;j} [1,3]  =  -{\frac{ {\hat x}_{k|k-1;i;j}^U} {{\hat r}_{k|k-1;i;j}^2}} \left(  {\frac{\partial {\hat r}_{k|k-1;i;j}}{\partial \gamma_{k-1;j}}} \right)} \\ & &  - \left[ {\frac{\partial y_{k-1;j}^U}{\partial \gamma_{k-1;j}}} \left( {\frac{1}{{\hat r}_{k|k-1;i;j}}} \right) - {\frac{y_{k-1;j}^U}{ {\hat r}_{k|k-1;i;j}^2}}  \left( {\frac{\partial  {\hat r}_{k|k-1;i;j} }{\partial \gamma_{k-1;j}}} \right) \right]   \\   \lefteqn{ {\hat {\bf H}}^{{\rm deriv},~\gamma}_{k;i;j} [2,1]  =    2 {\frac{ {\hat x}_{k|k-1;i;j}^N} {{\hat f}_{k|k-1;i;j}^3}} {\frac{\partial {\hat f}_{k|k-1;i;j}}{\partial \gamma_{k-1;j}}}} \\ & &  +   \left[ {\frac{\partial y_{k-1;j}^N}{\partial \gamma_{k-1;j}}} \left( {\frac{1}{{\hat f}_{k|k-1;i;j}^2}} \right) - 2 {\frac{y_{k-1;j}^N}{ {\hat f}_{k|k-1;i;j}^3}} {\frac{\partial  {\hat f}_{k|k-1;i;j} }{\partial \gamma_{k-1;j}}} \right]   \\   \lefteqn{  {\hat {\bf H}}^{{\rm deriv},~\gamma}_{k;i;j} [2,2]  =    -2 {\frac{ {\hat x}_{k|k-1;i;j}^E} {{\hat f}_{k|k-1;i;j}^3}} {\frac{\partial {\hat f}_{k|k-1;i;j}}{\partial \gamma_{k-1;j}}}} \\ & &  -   \left[ {\frac{\partial y_{k-1;j}^E}{\partial \gamma_{k-1;j}}} \left( {\frac{1}{{\hat f}_{k|k-1;i;j}^2}} \right) - 2 {\frac{y_{k-1;j}^E}{ {\hat f}_{k|k-1;i;j}^3}} {\frac{\partial  {\hat f}_{k|k-1;i;j} }{\partial \gamma_{k-1;j}}} \right]   \\   \lefteqn{  {\hat {\bf H}}^{{\rm deriv},~\gamma}_{k;i;j} [3,1]  =   -\left( 1 -  \left( {\frac{ {\hat \Delta}_{k|k-1;i;j}^U}{   {\hat r}_{k|k-1;i;j}}} \right)^2 \right)^{-{\frac{3}{2}}}} \\ & & \times {\frac{  \left( {\hat \Delta}_{k|k-1;i;j}^U \right)^3 {\hat \Delta}_{k|k-1;i;j}^E}{   {\hat r}_{k|k-1;i;j}^6}}{\frac{\partial  {\hat r}_{k|k-1;i;j} }{\partial \gamma_{k-1;j}}}    \\   &   &  -2 \left( 1 -  \left( {\frac{ {\hat \Delta}_{k|k-1;i;j}^U}{   {\hat r}_{k|k-1;i;j}}} \right)^2 \right)^{-{\frac{1}{2}}}  \\ & & \times  \left( {\frac{   {\hat \Delta}_{k|k-1;i;j}^U  {\hat \Delta}_{k|k-1;i;j}^E}{   {\hat r}_{k|k-1;i;j}^4}}{\frac{\partial  {\hat r}_{k|k-1;i;j} }{\partial \gamma_{k-1;j}}} \right. \\ & & \left.   - {\frac{1}{2}}  {\frac{   {\hat \Delta}_{k|k-1;i;j}^U }{   {\hat r}_{k|k-1;i;j}^2}} \cdot  \right.   \left[   -{\frac{{\hat x}_{k|k-1;i;j}^E}{ {\hat r}_{k|k-1;i;j}^2}}  {\frac{\partial  {\hat r}_{k|k-1;i;j} }{\partial \gamma_{k-1;j}}} - \right. \\ & &  \left. \left.  \left \{   {\frac{\partial y_{k-1;j}^E}{\partial \gamma_{k-1;j}}} {\frac{1}{ {\hat r}_{k|k-1;i;j}}} -  {\frac{y^E_{k-1;j}}{ {\hat r}_{k|k-1;i;j}^2}} {\frac{\partial  {\hat r}_{k|k-1;i;j} }{\partial \gamma_{k-1;j}}}      \right \}  \right]  \right)     \\   \lefteqn{ {\hat {\bf H}}^{{\rm deriv},~\gamma}_{k;i;j} [3,2]  =   -\left( 1 -  \left( {\frac{ {\hat \Delta}_{k|k-1;i;j}^U}{   {\hat r}_{k|k-1;i;j}}} \right)^2 \right)^{-{\frac{3}{2}}}} \\ & & \times {\frac{  \left( {\hat \Delta}_{k|k-1;i;j}^U \right)^3 {\hat \Delta}_{k|k-1;i;j}^N}{   {\hat r}_{k|k-1;i;j}^6}}{\frac{\partial  {\hat r}_{k|k-1;i;j} }{\partial \gamma_{k-1;j}}}  \\   &   &  -2 \left( 1 -  \left( {\frac{ {\hat \Delta}_{k|k-1;i;j}^U}{   {\hat r}_{k|k-1;i;j}}} \right)^2 \right)^{-{\frac{1}{2}}} \\ & & \times  \left( {\frac{  {\hat \Delta}_{k|k-1;i;j}^U  {\hat \Delta}_{k|k-1;i;j}^N}{   {\hat r}_{k|k-1;i;j}^4}}{\frac{\partial  {\hat r}_{k|k-1;i;j} }{\partial \gamma_{k-1;j}}}  \right. \\ & &  - {\frac{1}{2}}{\frac{   {\hat \Delta}_{k|k-1;i;j}^U }{   {\hat r}_{k|k-1;i;j}^2}} \cdot    \left[   -{\frac{{\hat x}_{k|k-1;i;j}^N}{ {\hat r}_{k|k-1;i;j}^2}}  {\frac{\partial  {\hat r}_{k|k-1;i;j} }{\partial \gamma_{k-1;j}}} -  \right.    \\ & & \left. \left.   \left \{   {\frac{\partial y_{k-1;j}^N}{\partial \gamma_{k-1;j}}} {\frac{1}{ {\hat r}_{k|k-1;i;j}}} -  {\frac{y^N_{k-1;j}}{ {\hat r}_{k|k-1;i;j}^2}} {\frac{\partial  {\hat r}_{k|k-1;i;j} }{\partial \gamma_{k-1;j}}}      \right \}  \right]  \right)\\   \lefteqn{  {\hat {\bf H}}^{{\rm deriv},~\gamma}_{k;i;j} [3,3]  =   \left( 1 -  \left( {\frac{ {\hat \Delta}_{k|k-1;i;j}^U}{   {\hat r}_{k|k-1;i;j}}} \right)^2 \right)^{-{\frac{3}{2}}}} \\ & & \times  {\frac{\partial  {\hat r}_{k|k-1;i;j} }{\partial \gamma_{k-1;j}}} \left(   {\frac{  \left( {\hat \Delta}_{k|k-1;i;j}^U \right)^2}{   {\hat r}_{k|k-1;i;j}^4}} - {\frac{  \left( {\hat \Delta}_{k|k-1;i;j}^U \right)^4 }{   {\hat r}_{k|k-1;i;j}^6}} \right)   \\   &   &  +  \left( 1 -  \left( {\frac{ {\hat \Delta}_{k|k-1;i;j}^U}{   {\hat r}_{k|k-1;i;j}}} \right)^2 \right)^{-{\frac{1}{2}}}  \left( {\frac{ 1}{   {\hat r}_{k|k-1;i;j}^2}}{\frac{\partial  {\hat r}_{k|k-1;i;j}}{\partial \gamma_{k-1;j}}}    \right. \\ & & \left.   -2 {\frac{  \left( {\hat \Delta}_{k|k-1;i;j}^U \right)^2 }{   {\hat r}_{k|k-1;i;j}^4}}{\frac{\partial  {\hat r}_{k|k-1;i;j} }{\partial \gamma_{k-1;j}}}   \right.  + {\frac{   {\hat \Delta}_{k|k-1;i;j}^U }{   {\hat r}_{k|k-1;i;j}^2}}      \\ & & \times    \left. \left[   -{\frac{{\hat x}_{k|k-1;i;j}^U}{ {\hat r}_{k|k-1;i;j}^2}}  {\frac{\partial  {\hat r}_{k|k-1;i;j} }{\partial \gamma_{k-1;j}}} - \left \{   {\frac{\partial y_{k-1;j}^U}{\partial \gamma_{k-1;j}}} {\frac{1}{ {\hat r}_{k|k-1;i;j}}}  \right. \right. \right. \\ & & \left.  \left. \left. -  {\frac{y^U_{k-1;j}}{ {\hat r}_{k|k-1;i;j}^2}} {\frac{\partial  {\hat r}_{k|k-1;i;j} }{\partial \gamma_{k-1;j}}}      \right \}  \right]  \right) \end{eqnarray*}   where    \begin{eqnarray*} {\frac{\partial y_{k-1;j}^E}{\partial \gamma_{k-1;j}}} & = & \left. {\frac{\partial y^E_{k-1;j}}{\partial \gamma}} \right|_{\gamma = \gamma_{k-1;j}} = -\sin \left( \gamma_{k-1;j} \right), \\ {\frac{\partial y_{k-1;j}^N}{\partial \gamma_{k-1;j}}} & = & \left. {\frac{\partial y^N_{k-1;j}}{\partial \gamma}} \right|_{\gamma = \gamma_{k-1;j}} = \cos \left( \gamma_{k-1;j} \right),    \\ {\frac{\partial y_{k-1;j}^U}{\partial \gamma_{k-1;j}}} & = & 0,    \\ {\hat f}_{k|k-1;i,j}  &  = &      \left[ \left(   {\hat \Delta}_{k|k-1;i;j}^E  \right)^2 + \left(   {\hat \Delta}_{k|k-1;i;j}^N  \right)^2  \right]^{{\frac{1}{2}}}, ~~{\rm and} \\    {\frac{\partial  {\hat f}_{k|k-1;i;j} }{\partial \gamma_{k-1;j}}} & = & \left.   {\frac{\partial f_{k|k-1;i;j}}{\partial \gamma}} \right|_{\gamma= \gamma_{k-1;j}} =  \left(  {\hat f}_{k|k-1;i;j} \right)^{-1} \times \\ & &  \left(  {\hat \Delta}_{k|k-1;i;j}^E \sin \left(  \gamma_{k-1;j}  \right) - \right. \\   & &  \left.   {\hat \Delta}_{k|k-1;i;j}^N \cos \left(  \gamma_{k-1;j}  \right)  \right).   \end{eqnarray*}

\subsection{Equation (\ref{eqn:uDeriv})}{\label{sctn:Appendix_uDeriv}}

The calculations for  Equation (\ref{eqn:uDeriv}) are very similar to those of Equation (\ref{eqn:gammaDeriv}), except instead of taking the derivative with respect to $\gamma$, the derivative is taken with respect to $y^U$.   The necessary derivatives are given below.

\begin{eqnarray*} \lefteqn{  \left.  {\frac{\partial {\hat {\pi }}_{k;i;j}^d \left( {\pmb \xi}_{k;j} \right)}{\partial y^U}} \right|_{y^U = y^U_{k-1;j}}} \\  & = & -{\hat {\pi }}_{k;i;j}^d \left( {\pmb \xi}_{k;j} \right) \cdot {\frac{ {\hat r}_{k|k-1;i;j}}{50}} \left( {\frac{\partial {\hat r}_{k|k-1;i;j}}{\partial y^U_{k-1;j}}} \right), ~~{\rm and} \\  \lefteqn{{\frac{\partial  {\hat r}_{k|k-1;i;j} }{\partial y^U_{k-1;j}}}} \\  & = & \left.   {\frac{\partial r_{k|k-1;i;j}}{\partial y^U}} \right|_{y^U= y^U_{k-1;j}} =    \left. - {\hat \Delta}_{k|k-1;i;j}^U \right/ {\hat r}_{k|k-1;i;j}  .\end{eqnarray*}  

\noindent Below are the elements of   $ \left. {\frac{\partial {\hat {\bf H }}_{k;i;j} \left( {\pmb \xi}_{k;j} \right)}{\partial y^U}}  \right|_{\gamma = \gamma_{k-1;j}, y^U = y^U_{k-1;j} }.$ For notational simplicity, we denote the $(a,b)^{\rm th}$ element in this matrix as  ${\hat {\bf H}}^{{\rm deriv},~y^U}_{k;i;j} [a,b],$ and unless otherwise stated, ${\hat {\bf H}}^{{\rm deriv},~\gamma}_{k;i;j} [a,b] = 0.$       \begin{eqnarray*} {\hat {\bf H}}^{{\rm Deriv},~y^U}_{k;i;j} [1,1] & = & -{\frac{ {\hat x}_{k|k-1;i;j}^E} {{\hat r}_{k|k-1;i;j}^2}} {\frac{\partial {\hat r}_{k|k-1;i;j}}{\partial y^U_{k-1;j}}}    \\      {\hat {\bf H}}^{{\rm Deriv},~y^U}_{k;i;j} [1,2] & = & -{\frac{ {\hat x}_{k|k-1;i;j}^N} {{\hat r}_{k|k-1;i;j}^2}} {\frac{\partial {\hat r}_{k|k-1;i;j}}{\partial y^U_{k-1;j}}}     \\      {\hat {\bf H}}^{{\rm Deriv},~y^U}_{k;i;j} [1,3] & = & -{\frac{ {\hat x}_{k|k-1;i;j}^U} {{\hat r}_{k|k-1;i;j}^2}} {\frac{\partial {\hat r}_{k|k-1;i;j}}{\partial y^U_{k-1;j}}} - {\frac{1}{{\hat r}_{k|k-1;i;j}}}  \end{eqnarray*}     \begin{eqnarray*}   \lefteqn{ {\hat {\bf H}}^{{\rm Deriv},~y^U}_{k;i;j} [3,1]  =  \left( 1 -  \left( {\frac{ {\hat \Delta}_{k|k-1;i;j}^U}{   {\hat r}_{k|k-1;i;j}}} \right)^2 \right)^{-{\frac{3}{2}}}} \times \\ & &    {\frac{ \left(  {\hat \Delta}_{k|k-1;i;j}^U \right)^2  {\hat \Delta}_{k|k-1;i;j}^E}{   {\hat r}_{k|k-1;i;j}^4}}  \left[ -{\frac{ {\hat x}_{k|k-1;i;j}^U} {{\hat r}_{k|k-1;i;j}^2}} {\frac{\partial {\hat r}_{k|k-1;i;j}}{\partial y^U_{k-1;j}}} -  \right. \\ & &  \left.     \left( {\frac{1}{{\hat r}_{k|k-1;i;j}}} -{\frac{ y_{k-1;i;j}^U} {{\hat r}_{k|k-1;i;j}^2}} {\frac{\partial {\hat r}_{k|k-1;i;j}}{\partial y^U_{k-1;j}}}  \right) \right] - \\ & &   \left( 1 -  \left( {\frac{ {\hat \Delta}_{k|k-1;i;j}^U}{   {\hat r}_{k|k-1;i;j}}} \right)^2 \right)^{-{\frac{1}{2}}}     \left(    {\frac{ \left(  {\hat \Delta}_{k|k-1;i;j}^U  {\hat \Delta}_{k|k-1;i;j}^E  \right)}{   {\hat r}_{k|k-1;i;j}^4}}  \right. \\ & & \cdot  {\frac{\partial {\hat r}_{k|k-1;i;j}}{\partial y^U_{k-1;j}}}              - {\frac{ \left(  {\hat \Delta}_{k|k-1;i;j}^E \right)}{   {\hat r}_{k|k-1;i;j}}} \left(-2 {\frac{ {\hat x}_{k|k-1;i;j}^U} {{\hat r}_{k|k-1;i;j}^3}}  \right. \\ & &  \left. \left.  {\frac{\partial {\hat r}_{k|k-1;i;j}}{\partial y^U_{k-1;j}}} - \left[  {\frac{1}{{\hat r}_{k|k-1;i;j}^2}} - 2{\frac{ {\hat y}_{k-1;i;j}^U} {{\hat r}_{k|k-1;i;j}^3}} {\frac{\partial {\hat r}_{k|k-1;i;j}}{\partial y^U_{k-1;j}}}   \right]  \right) \right)     \\   \lefteqn{    {\hat {\bf H}}^{{\rm Deriv},~y^U}_{k;i;j} [3,2]  =  \left( 1 -  \left( {\frac{ {\hat \Delta}_{k|k-1;i;j}^U}{   {\hat r}_{k|k-1;i;j}}} \right)^2 \right)^{-{\frac{3}{2}}} } \\   & & {\frac{ \left(  {\hat \Delta}_{k|k-1;i;j}^U \right)^2  {\hat \Delta}_{k|k-1;i;j}^N}{   {\hat r}_{k|k-1;i;j}^4}}  \left[ -{\frac{ {\hat x}_{k|k-1;i;j}^U} {{\hat r}_{k|k-1;i;j}^2}} {\frac{\partial {\hat r}_{k|k-1;i;j}}{\partial y^U_{k-1;j}}} -  \right.    \\ & &  \left. \left( {\frac{1}{{\hat r}_{k|k-1;i;j}}} -{\frac{ {y}_{k-1;i;j}^U} {{\hat r}_{k|k-1;i;j}^2}} {\frac{\partial {\hat r}_{k|k-1;i;j}}{\partial y^U_{k-1;j}}}  \right) \right] -    \\ & &   \left( 1 -  \left( {\frac{ {\hat \Delta}_{k|k-1;i;j}^U}{   {\hat r}_{k|k-1;i;j}}} \right)^2 \right)^{-{\frac{1}{2}}}     \left(    {\frac{ \left(  {\hat \Delta}_{k|k-1;i;j}^U  {\hat \Delta}_{k|k-1;i;j}^N  \right)}{   {\hat r}_{k|k-1;i;j}^4}}   \right. \\ & &  \cdot {\frac{\partial {\hat r}_{k|k-1;i;j}}{\partial y^U_{k-1;j}}}       - {\frac{ \left(  {\hat \Delta}_{k|k-1;i;j}^N \right)}{   {\hat r}_{k|k-1;i;j}}} \left(-2 {\frac{ {\hat x}_{k|k-1;i;j}^U} {{\hat r}_{k|k-1;i;j}^3} } \cdot  \right. \\ & & \left. \left.  {\frac{\partial {\hat r}_{k|k-1;i;j}}{\partial y^U_{k-1;j}}} - \left[  {\frac{1}{{\hat r}_{k|k-1;i;j}^2}} - 2{\frac{ {\hat y}_{k-1;i;j}^U} {{\hat r}_{k|k-1;i;j}^3}} {\frac{\partial {\hat r}_{k|k-1;i;j}}{\partial y^U_{k-1;j}}}   \right]  \right) \right)    \\   \lefteqn{ {\hat {\bf H}}^{{\rm Deriv},~y^U}_{k;i;j} [3,3]  =   \left[ 1 -  \left( \xiRatio \right)^2 \right]^{-\frac{3}{2}}} \\ & &  \left[ {\frac{1}{\rHat}} - {\frac{ \left( \DeltaUp \right)^2}{\rHat^3}} \right]   \left( \xiRatio \right) \cdot \\ & & \left[     - {\frac{{\hat x}^U_{k|k-1;i;j}}{ {\hat r}_{k|k-1;i;j}^2}} {\frac{\partial {\hat r}_{k|k-1}}{\partial y^U_{k-1;j}}} -  \left( {\frac{1}{ {\hat r}_{k|k-1;i;j}}} + \right. \right. \\ & & \left. \left. {\frac{y^U_{k-1;j}}{ {\hat r}_{k|k-1;i;j}^2}}  {\frac{\partial {\hat r}_{k|k-1;i;j}}{\partial y^U_{k-1;j}}}   \right) \right]  - \left[ 1 - \left( \xiRatio \right)^2 \right]^{-{\frac{1}{2}}} \cdot   \\ & &  \times \left \{ - {\frac{1}{ \rHat^2}} { {\frac{\partial \rHat}{\partial y^U_{k-1;j}}}}  - \right.  \left[ \psiRatio  \right.  \cdot \\ & & \left(  - {\frac{{\hat x}^U_{k|k-1;i;j}}{ {\hat r}_{k|k-1;i;j}^2}} {\frac{\partial {\hat r}_{k|k-1}}{\partial y_{k-1;j}^U}} -  \left( {\frac{1}{ {\hat r}_{k|k-1;i;j}}} + \right. \right.   \\& &  \left. \left. {\frac{y^U_{k-1;j}}{ {\hat r}_{k|k-1;i;j}^2}} {\frac{\partial {\hat r}_{k|k-1;i;j}}{\partial y^u_{k-1;j}}} \right) \right) +   \left( \betaZRatio \right) \cdot \\ & &   \left(  - 2 {\frac{{\hat x}^U_{k|k-1;i;j}}{ {\hat r}_{k|k-1;i;j}}} {\frac{\partial {\hat r}_{k|k-1}}{\partial y_{k-1;j}^U}} - \left( {\frac{1}{{\hat r}_{k|k-1;i;j}^2}} - \right. \right. \\ & & \left.  \left. \left. \left.  2 {\frac{y^U_{k-1;j}}{ {\hat r}_{k|k-1;i;j}^3}} {\frac{\partial {\hat r}_{k|k-1;i;j}}{\partial y^U_{k-1;j}}} \right)    \right) \right] \right \}     \end{eqnarray*}

\subsection{Comparing Communication Probabilities}{\label{sctn:CmnctProb}}

\begin{figure}[th]
   \begin{center}
   \begin{tabular}{c}
      \includegraphics[trim = 1cm 7cm 1cm 7cm, scale = 0.45]{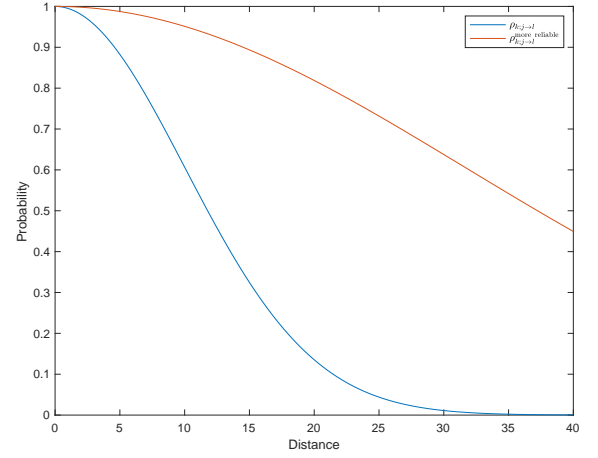}
   \end{tabular}
   \end{center}
\caption{The probability of communication between agents, $p_{k;j \rightarrow l}$ and $p_{k;j \rightarrow l}^{\text {more~reliable}}.$}
   \label{fig:comparingProbComm} 
\end{figure}

\begin{figure}[th]
   \begin{center}
   \begin{tabular}{c}
      \includegraphics[trim = 1cm 7cm 1cm 7cm, scale = 0.45]{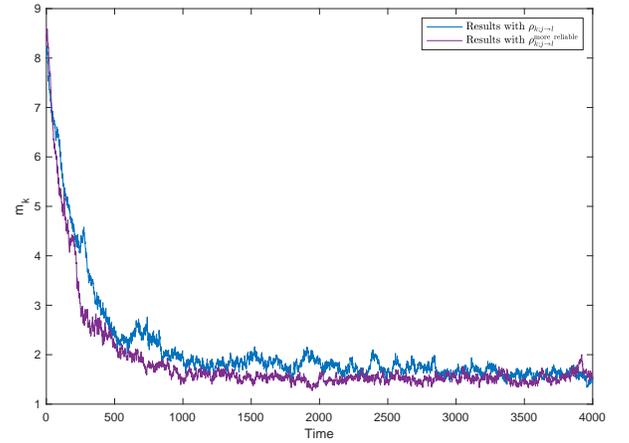}
   \end{tabular}
   \end{center}
\caption{Median distance (across 100 simulations)  from target to closest agent over time assuming two different communication models. The total number of targets is 2, and the total number of agents is 2.}
   \label{fig:comparingConvergence} 
\end{figure}

The communication probability used in the simulations of this study is given Equation (\ref{eqn:communic_Prob}). We referred to this communication probability in the text as ``non-ideal."  In this subsection of the Appendix, we show that if the communication probability between the agents were better, the performance of our algorithm would improve. We specifically compare the performance of our algorithm when the communication probability between agents takes the form given in Equation (\ref{eqn:communic_Prob})  to the performance of our algorithm when the communication probability between agents takes the form \begin{eqnarray*}  \rho_{k;j \rightarrow l}^{\text {more~reliable}} &=  & \exp \left \{ -  \left. \left[ \left( y_{k;j}^E - y_{k;l}^E \right)^2   +   \left( y_{k;j}^N - y_{k;l}^N \right)^2    \right. \right. \right. \\ & & \left. \left. \left. ~~~~~~~~~~+ \left( y_{k;j}^U - y_{k;l}^U \right)^2 \right] \right/2000 \right \}.   \end{eqnarray*} The plot in Figure \ref{fig:comparingProbComm} illustrates the difference between these two probabilities, and the plot in Figure \ref{fig:comparingConvergence} shows the results of two simulation studies meant to compare the two communication models. The simulation studies are identical to the ones illustrated in Figures \ref{fig:medianDist_T2} and \ref{fig:medianDist_T4}; in each study, one hundred simulations of 4000 time steps were done, and the median minimum distance to an agent from each target was calculated at each time point. We denote this distance at time $k$ as $m_k$, and in these simulations, there were two agents and two targets ($A = 2$ and $T = 2$). The results of the simulation study using the communication probability $\rho_{k;j \rightarrow l}$ are in blue, and those using the communication probability $\rho_{k;j \rightarrow l}^{\text {more~reliable}}$ are in magenta. It is clear that when communications are better, the agents converge more 	quickly and more closely to the targets.

\end{document}